%% file: acl_latex.tex
\documentclass[11pt]{article}

\usepackage[preprint]{acl}

\usepackage{times}
\usepackage{latexsym}

\usepackage{enumitem}
\usepackage{hyperref}
\usepackage{xcolor}
\usepackage{natbib}
\usepackage{graphicx}
\usepackage{colortbl}
\usepackage{rotating}
\usepackage{tcolorbox}
\usepackage{subcaption}
\usepackage{dashrule}
\usepackage{amsmath}
\usepackage{makecell}
\usepackage{pifont}
\usepackage{xcolor}
\usepackage{booktabs}

\newcommand{\Tick}{{\color{teal}\ding{51}}}
\newcommand{\Cross}{{\color{red}\ding{55}}}
\newcommand{\CtoC}{\Tick$\rightarrow$\Tick}
\newcommand{\CtoW}{\Tick$\rightarrow$\Cross}
\newcommand{\WtoC}{\Cross$\rightarrow$\Tick}
\newcommand{\WtoW}{\Cross$\rightarrow$\Cross}

\newcommand{\alberto}[1]{\textcolor{black}{#1}}

\newcommand{\cut}[1]{}
\newcommand{\old}[1]{}

\usepackage[T1]{fontenc}

\usepackage[utf8]{inputenc}

\usepackage{microtype}

\usepackage{inconsolata}

\usepackage{graphicx}

\title{Uncertainty Is Not a Safety Net for Clinical VQA, but Can It Anticipate Model Failure?}

\author{
 \textbf{Arnisa Fazla\textsuperscript{1,2\thanks{These authors contributed equally to this work.}}},
 \textbf{Alberto Testoni\textsuperscript{1,2\textsuperscript{$\ast$}}},
 \textbf{Ameen Abu-Hanna\textsuperscript{1,2}},\\
 \textbf{Barbara Plank\textsuperscript{3,4}},
\textbf{Iacer Calixto\textsuperscript{1,2}}
\\
 \textsuperscript{1}Department of Medical Informatics, Amsterdam University Medical Center, 
 \\ University of Amsterdam, Amsterdam, The Netherlands.
\\ \textsuperscript{2}Amsterdam Public Health, Methodology, Amsterdam, The Netherlands.
\\
\textsuperscript{3}MaiNLP, Center for Information and Language Processing, LMU Munich, Germany.\\
\textsuperscript{4}Munich Center for Machine Learning (MCML), Munich, Germany.
 \\
\small{
\textbf{Correspondence:} \href{mailto:a.fazla@amsterdamumc.nl}{a.fazla@amsterdamumc.nl}
}
 }

\begin{document}
\maketitle
\begin{abstract}

Safe deployment of clinical vision-language models (VLMs) requires reliable uncertainty estimation (UE): a signal indicating when predictions should be trusted or escalated to a clinician. We test whether current UE methods actually deliver this signal. Benchmarking 8 methods across 12 VLMs on clinical visual question-answering (VQA), we find that UE quality is not an intrinsic property of the UE method: it tracks model accuracy, degrading precisely where the model performance is weakest, and therefore where reliability is most needed. When we stress-test models by hiding the correct option among the multiple-choice answers (NOTA perturbations), accuracy collapses while uncertainty barely changes, leaving models systematically miscalibrated. Yet, we find that uncertainty on the unperturbed input reliably anticipates which predictions will collapse under NOTA, indicating that UE in current VLMs carries diagnostic information about model fragility. Our results position UE as a diagnostic tool for identifying fragile predictions and motivate perturbation-based evaluation as a path toward safe clinical deployment.

\end{abstract}

\section{Introduction}
\label{sec:intro}

\begin{figure}[t]
    \begin{minipage}{1\columnwidth}
    \centering
    \includegraphics[width=1\linewidth]
    {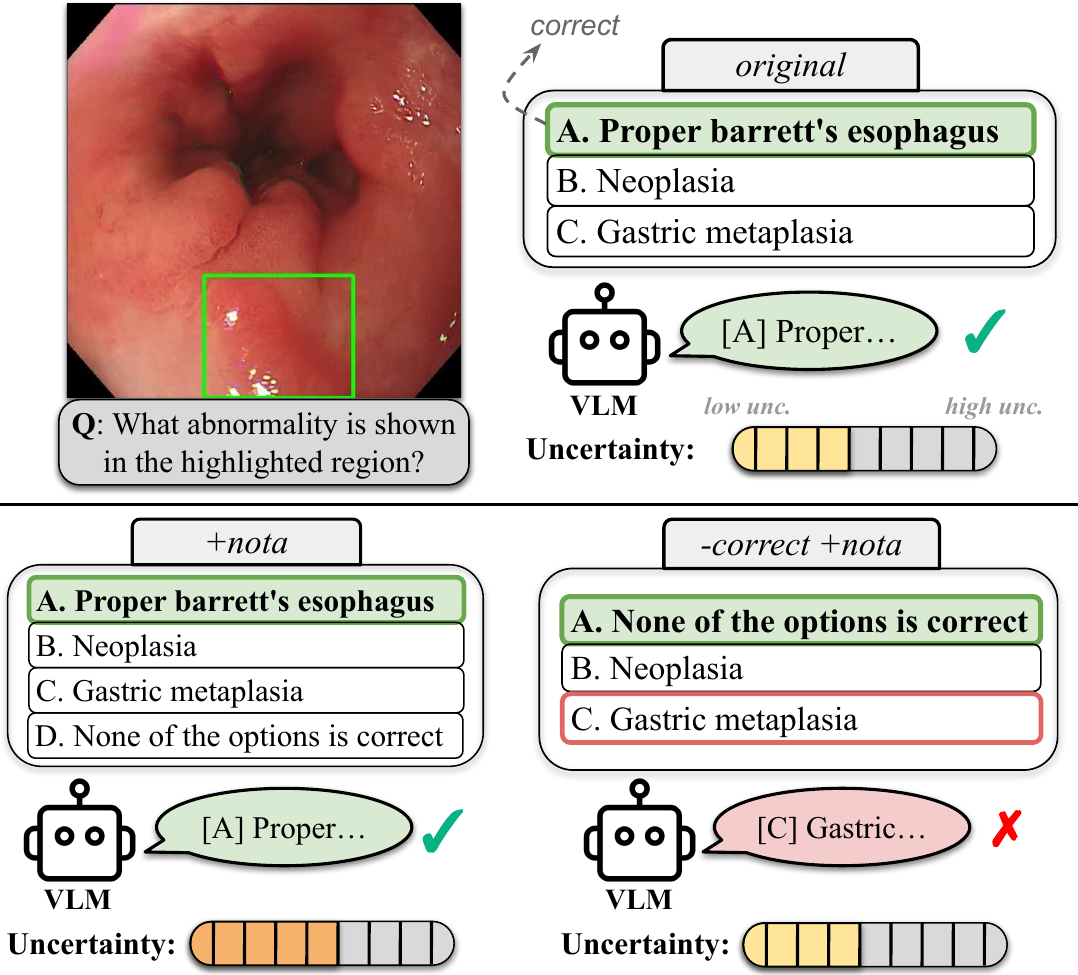}
    \caption{Illustration of NOTA perturbations. Starting from the original question (top), we construct two variants: \textit{+nota} adds a misleading \textit{None of the options is correct} option; \textit{ -correct+nota} replaces the correct answer (in green) with the NOTA option. Under the \mbox{\textit{-correct+nota}} option, the model becomes incorrect while its uncertainty barely changes, mirroring systematic patterns we observe across models and methods.
    }
    \label{fig:dataset_example}
    \end{minipage}
\end{figure}

Clinical workflows rely on multimodal information, including medical images, clinical notes, and textual reports. Vision-language models (VLMs) can jointly process visual and textual data, enabling applications such as diagnosis support, severity grading, and automated report generation \citep{9999689, 9745795}. 
\alberto{Yet deploying VLMs in clinical practice is constrained by a well-documented failure mode: they can produce fluent, confident answers that are wrong, with no internal signal distinguishing the two \citep{groot-valdenegro-toro-2024-overconfidence}. Without such a signal, errors propagate silently into clinical decisions. Uncertainty estimation (UE) addresses this problem by quantifying how much each prediction can be trusted, enabling clinicians to abstain, defer, or escalate cases where the model is unreliable \citep{DAWOOD2023102861}.}

This promise, however, depends on UE methods themselves being reliable. Most evaluations of UE are conducted in controlled settings: fixed input distributions, well-formed questions, and answer spaces guaranteed to contain a correct option. Clinical practice often violates all of these assumptions. Models encounter inputs spanning diverse imaging modalities, clinical specialties, and task types \citep{LIN2023102611}, and may be asked questions that are ambiguous, unanswerable, or whose correct answer is not among the options provided. Because exhaustively testing every such scenario is infeasible, a practical safety strategy requires UE itself to flag the cases where its own predictions cannot be trusted. Whether current UE methods meet this bar is an open question.

Despite growing interest in uncertainty estimation for VLMs, systematic evaluation in clinical VQA remains limited. While most prior clinical UE benchmarks are text-only \citep{10.1093/jamia/ocae254, testoni-calixto-2026-mind}, existing studies evaluating UE in VLMs focus on non-clinical vision-language benchmarks \citep{kostumov2024uncertaintyawareevaluationvisionlanguagemodels, zhang2024vluncertaintydetectinghallucinationlarge}, and evaluate a few baseline UE methods like verbalized confidence \citep{groot-valdenegro-toro-2024-overconfidence} and token-level probability \citep{info:doi/10.2196/64348,ICLR2025_21b5788d}. To our knowledge, there exists no benchmark that systematically evaluates UE methods on clinical VQA, nor analyzes their behavior across relevant dimensions such as imaging modalities, clinical specialties, and other clinical factors.

We address this gap with a systematic evaluation of post-hoc UE in clinical VQA. We benchmark 8 UE methods, spanning \textit{logit-based}, \textit{consistency-based}, and recent \textit{embedding-based} methods, across 12 VLMs (2 biomedical and 10 general-purpose, including a proprietary model) from 5 model families and multiple sizes, measuring both calibration (alignment between confidence and accuracy) and discrimination (ability to distinguish correct from incorrect predictions). We adopt multiple-choice question answering (MCQA) as a controlled testbed, following standard practice in clinical UE evaluation \citep{singhal2023large,NEURIPS2024_fde7f40f, 10.1093/jamia/ocae254, testoni-calixto-2026-mind}: while not capturing the full complexity of real-world clinical decision-making, MCQA offers a clear ground truth and discrete answer space, allowing clean comparison of UE performance across settings without confounders from answer evaluation. Beyond overall performance, we stratify results across imaging modalities, and evaluate \textit{None of the Above} (NOTA) perturbations to probe robustness under input variations where the correct answer is not among the provided options \citep{griot2025large}.

We report three main findings. First, no single UE method dominates: logit-, consistency-, and embedding-based approaches each excel on different axes, and the strongest methods on average are not the strongest within any given clinical context. Second, the dominant predictor of UE quality is not the method but rather the underlying model's accuracy on the input: discrimination and calibration co-vary with accuracy across imaging modalities, with degradation concentrated precisely where UE would be most clinically useful. Third, the NOTA perturbations expose a decoupling of uncertainty from accuracy: accuracy collapses when the correct answer is removed, but uncertainty does not rise to match, even for methods that appear well-calibrated under standard evaluation. 

A complementary finding reframes this picture: uncertainty estimated on the original (unperturbed) input carries predictive signal about which predictions will collapse under NOTA, suggesting that UE retains diagnostic value even where its safety-signal interpretation breaks down. Together, our findings reveal that UE in clinical VLMs is neither uniformly reliable nor broken: its failures are systematic, its successes contextual, and its diagnostic value goes beyond the per-prediction calibration scores typically reported. We argue for evaluation protocols that surface this complexity before deployment. Our code is available at \href{https://github.com/arnisafazla/uncertainty-safety-net-clinical-vqa}{https://github.com/arnisafazla/uncertainty-safety-net-clinical-vqa}.

\section{Related Work}
\label{sec:related_work}

Large language models (LLMs) have been widely evaluated on text-only clinical knowledge tasks using multiple-choice question answering (MCQA) benchmarks such as MedQA, MedMCQA \citep{pmlr-v174-pal22a}, and PubMedQA \citep{jin-etal-2019-pubmedqa,singhal2023large}. Following the MCQA format, numerous studies have assessed model accuracy across individual clinical specialties, including pediatrics \citep{10.1001/jamapediatrics.2023.5750}, oncology \citep{doi:10.1056/AIoa2300151}, ophthalmology \citep{10.1001/jamaophthalmol.2023.1144}, radiology \citep{doi:10.1148/radiol.230987,info:doi/10.2196/64284}, and plastic surgery \citep{10.1093/asj/sjad130}. 

More comprehensive benchmarks have since emerged, extending beyond text-only QA to visual multiple-choice QA, including MedExQA \citep{kim-etal-2024-medexqa} and MedXpertQA \citep{pmlr-v267-zuo25a}. In parallel, vision–language models (VLMs) are increasingly evaluated using clinical MCQA and VQA benchmarks covering specific domains such as pediatrics \citep{bahaj2025pediatricsmqamultimodalpediatricsquestion} and gastroenterology \citep{Safavi_Naini_2025}, as well as broader multi-domain frameworks including MultiMedEval \citep{pmlr-v250-royer24a}, Asclepius \citep{liu-etal-2025-asclepius}, and GMAI-MMBench, which spans 284 datasets across 38 imaging modalities \citep{NEURIPS2024_ab7e02fd}. Collectively, these efforts provide extensive accuracy benchmarks across specialties and modalities, but do not examine uncertainty estimation.

Previous work on uncertainty estimation (UE) in clinical QA has largely focused on text-based tasks. \citet{testoni-calixto-2026-mind} benchmarked UE methods for text-based clinical MCQA across LLMs, analyzing calibration and discrimination across specialties and question types. Similarly, \citet{10.1093/jamia/ocae254} evaluated LLM calibration on diagnosis and treatment selection tasks. To assess UE robustness, \citet{testoni-calixto-2026-calibrated} evaluated how patient identity markers, including sexual orientation and religious affiliation, affect model accuracy and calibration. For VLMs, \citet{kostumov2024uncertaintyawareevaluationvisionlanguagemodels} assessed conformal prediction for uncertainty estimation in general-purpose VQA. \citet{zhang2024vluncertaintydetectinghallucinationlarge} evaluated hallucination detection under visual-textual perturbations using semantic entropy. 
\citet{NEURIPS2024_fde7f40f} evaluated VLMs on MedVQA tasks and estimated confidence by prompting models with “Are you sure you accurately answered the question?” and using the probability of the “yes” token. Their evaluation relied on simple metrics such as uncertainty-based accuracy and overconfidence ratio. 

Within uncertainty estimation for clinical MCQA, only a few works have examined the role of a \textit{None of the Above} (NOTA) option. \citet{kadavath2022languagemodelsmostlyknow} found that introducing NOTA generally harms performance and calibration in text-only MCQA tasks, but did not distinguish between replacing correct versus incorrect answers, nor evaluate finer-grained uncertainty metrics. \citet{griot2025large} proposed MetaMedQA, a text-only medical MCQA dataset for unanswerable questions. They used self-reported confidence and coarse evaluation metrics but did not systematically evaluate uncertainty estimation (UE) methods. In contrast, our work evaluates a broad set of VLMs using diverse UE methods, providing a more detailed understanding of how NOTA perturbations affect both model predictions and uncertainty estimation.

\section{Methodology}
\label{sec:methodology}

\subsection{Task and dataset}
\label{subsec:dataset}
We use GMAI-MMBench \citep{NEURIPS2024_ab7e02fd}, a multiple-choice clinical VQA dataset. Each sample includes a medical image, a corresponding question, and 2-5 answer options (example in Figure~\ref{fig:dataset_example}). 

We extract an evaluation suite with 210 samples from each of the 8 most common imaging modalities (\textit{CT}, \textit{MRI}, \textit{Endoscopy}, \textit{Histopathology}, \textit{Fundus Photography}, \textit{X-ray}, \textit{Microscopy}, and \textit{Dermoscopy}) from the validation set, resulting in 1,680 samples in total.

\subsection{Models and answer extraction}
\label{subsec:models}
We evaluate 11 open-source instruction-tuned VLMs, including 9 \textbf{general-purpose} models: \texttt{LLaVA-v1.6-7B}, \texttt{LLaVA-v1.6-34B} \citep{liu2024llavanext}, \texttt{LLaVA-OV-8B} \citep{an2025llavaonevision15fullyopenframework}, \texttt{Qwen2-VL-7B} \citep{wang2024qwen2vlenhancingvisionlanguagemodels}, \texttt{Qwen2.5-VL-7B}, \texttt{Qwen2.5-VL-32B}, \texttt{Qwen2.5-VL-72B} \citep{bai2025qwen25vltechnicalreport}, \texttt{Molmo-7B}, \texttt{Molmo-72B} \citep{11092508}; and 2 
\textbf{biomedical} models:  \texttt{MedGemma-4B}, \texttt{MedGemma-27B} \citep{sellergren2026medgemmatechnicalreport}. In addition, we include one proprietary model, \texttt{GPT-4.1-Mini} \citep{openai_gpt4_1_mini_2025}. Model details, including access and licenses, are provided in Appendix~\ref{app_sec:models}.

During inference, models are prompted to return a concise free-form answer including the selected option in square brackets (e.g., [C]). Following \citet{testoni-calixto-2026-mind}, we use regular expressions to extract answers when the output matches this format. For the remaining cases ($\leq4.8\%$ of samples), we use \texttt{LLaMA-3.1-8B-Instruct} \citep{meta2024introducing} to extract the predicted option from the model response. Manual validation on 130 samples shows a 95\% extraction accuracy for this LLM-based extraction. Details on prompts, extraction, and statistics are provided in Appendix~\ref{app_sec:prompts}.

\subsection{Uncertainty estimation}
\label{subsec:UE}

We evaluate eight post-hoc UE methods spanning three families: logit-based, consistency-based, and embedding-based. This is broader than prior benchmarks, which typically focus on a narrow set of baselines such as verbalized confidence \citep{groot-valdenegro-toro-2024-overconfidence} or token-level probabilities \citep{info:doi/10.2196/64348,ICLR2025_21b5788d}, and enables comparison of diverse methods on the same data and models. The three families also span a practical trade-off: logit-based methods require a single inference pass but assume access to token probabilities; consistency- and embedding-based methods are black-box but require multiple generations, with the latter additionally relying on an external encoder. We describe each family below; formal definitions are provided in Appendix~\ref{app_subsec:definitions}.

\paragraph{Logit-based} Logit-based UE approaches estimate the confidence of the generated responses using the probabilities of the generated tokens. In this category, we evaluate \emph{Average Negative Log-Likelihood} (ANLL) and \emph{Max Negative Log-Likelihood} (Max NLL) \citep{manakul-etal-2023-selfcheckgpt}, which compute the average and maximum (respectively) of the negative log-likelihoods (NLL) of the output tokens as an efficient uncertainty proxy. We also introduce \emph{Label Negative Log-Likelihood} (Label NLL), a simple method that computes the NLL of the token corresponding to the selected answer choice (e.g., the token probability of ``[A]''). 

\paragraph{Consistency-based} Consistency-based UE approaches estimate uncertainty by measuring answer agreement across multiple stochastic decoding runs. \emph{Sample Consistency} (SC) \citep{10.1093/jamia/ocae254} computes uncertainty as $1 -$ the proportion of sampled outputs matching the majority-vote answer. \emph{Semantic Entropy} (SE) \citep{kuhn2023semantic,farquhar2024detecting} groups sampled responses by semantic equivalence and computes the entropy over the resulting clusters. In the MCQA setting, we cluster by the predicted answer option (see Section \ref{subsec:models} for the answer extraction method).

\paragraph{Embedding-based} Embedding-based UE approaches generate multiple outputs, embed each with an external model, and compute geometric features (e.g., variance, pairwise distances) of those points. We evaluate \emph{EigenEmbed} (EE) \citep{nguyen2026distanceneedradialdispersion}, which uses an external encoder to map model outputs to an embedding space and computes the trace of their covariance matrix. We also evaluate the \emph{Radial Dispersion Score} (RDS) \citep{nguyen2026distanceneedradialdispersion}, which measures the radial dispersion of embeddings from their centroid. For both methods, we use a biomedical encoder,\footnote{\href{https://huggingface.co/pritamdeka/BioBERT-mnli-snli-scinli-scitail-mednli-stsb}{BioBERT-mnli-snli-scinli-scitail-mednli-stsb}} which outperforms the encoder used by \citet{nguyen2026distanceneedradialdispersion} on our dataset (see Appendix~\ref{app_subsec:biomedical_comparison}).

\paragraph{Hybrid} \emph{PRobability Only} (PRO) \citep{Nguyen_Gupta_Le_2026} is a hybrid approach that requires both output logits and sampling multiple answers. It estimates the uncertainty as the entropy of the probabilities of sampled responses. 

\subsection{Evaluation}
\label{subsec:evaluation}

Following \citet{manakul-etal-2023-selfcheckgpt}, we generate 10 responses per sample using nucleus sampling with $p=0.9$ and temperature $T=0.6$, consistent with \citet{testoni-calixto-2026-mind}. We apply this multi-generation setting uniformly across all UE methods, including those that do not strictly require it, to ensure a fair comparison. To mitigate position bias, we shuffle each question's answer options up to four times, using all unique orders when fewer than four exist.
We compute accuracy by majority vote against the ground-truth label, resolving ties randomly. We evaluate UE along two complementary axes: \emph{discrimination}, the ability of the uncertainty estimates to separate correct from incorrect predictions, measured by AUROC (higher is better); and \emph{calibration} (lower is better), the alignment between confidence and accuracy, measured by Expected Calibration Error (ECE, 10 bins, following \citealp{rivera-etal-2024-combining}) and the Brier Score \citep{glenn1950verification}. For statistical comparisons, we use the McNemar test for accuracy changes, paired bootstrap tests (2,000 bootstraps) for within-subset comparisons, and unpaired bootstrap tests (2,000 bootstraps) for cross-subset uncertainty comparisons, all at $p \leq 0.05$.

\subsection{NOTA Experiments}
\label{subsec:nota_methodology}

\input{main_table}

Standard MCQA evaluation assumes the correct answer is always among the options. Clinical practice routinely violates this assumption: questions can be ambiguous, mis-specified, or have no correct listed answer. MCQA accuracy is also known to shift under benign format changes \citep{rosenthal2025unexploredflawsmultiplechoicevqa}, so calibration measured in the standard setting (unperturbed inputs, correct answer present) may not transfer. We probe this with a controlled \textit{None of the Above} (NOTA) stress test (Figure~\ref{fig:dataset_example} for an example, with results presented in Section \ref{sec:NOTA}). 

We modify each question in two ways: (1) adding a \textit{None of the options is correct} choice while keeping the correct answer present (\texttt{+nota}), mimicking an escape option as a distractor; and (2) replacing the correct answer with the NOTA option (\texttt{-correct+nota}), so that NOTA itself becomes the correct choice. The two variants isolate distinct failure modes: distraction by a plausible wrong option, and failure to recognize when no listed option is correct.

These perturbations let us test two hypotheses about UE. \emph{H1} (robustness under perturbation): under both NOTA settings, average uncertainty rises as accuracy drops. This is the core property assumed whenever UE is used as a threshold for abstention or escalation, yet prior work has evaluated only the robustness of accuracy, not of UE itself \citep{griot2025large, kadavath2022languagemodelsmostlyknow}. \emph{H2} (predictive signal): under both NOTA settings, the model is more likely to flip its answer for samples that initially had higher uncertainty. A positive result would establish a novel use of UE as a response-fragility signal: rather than merely indicating how confident the model is in a given answer, UE could identify answers that are unlikely to remain stable under small perturbations. Such a signal could support abstention or escalation by flagging predictions that may be too fragile to surface without further review.

\section{Benchmarking Results}
\label{sec:results}

\subsection{Performance across models and methods}
\label{subsec:results_main}

\begin{figure*}
\centering
\includegraphics[width=\textwidth]{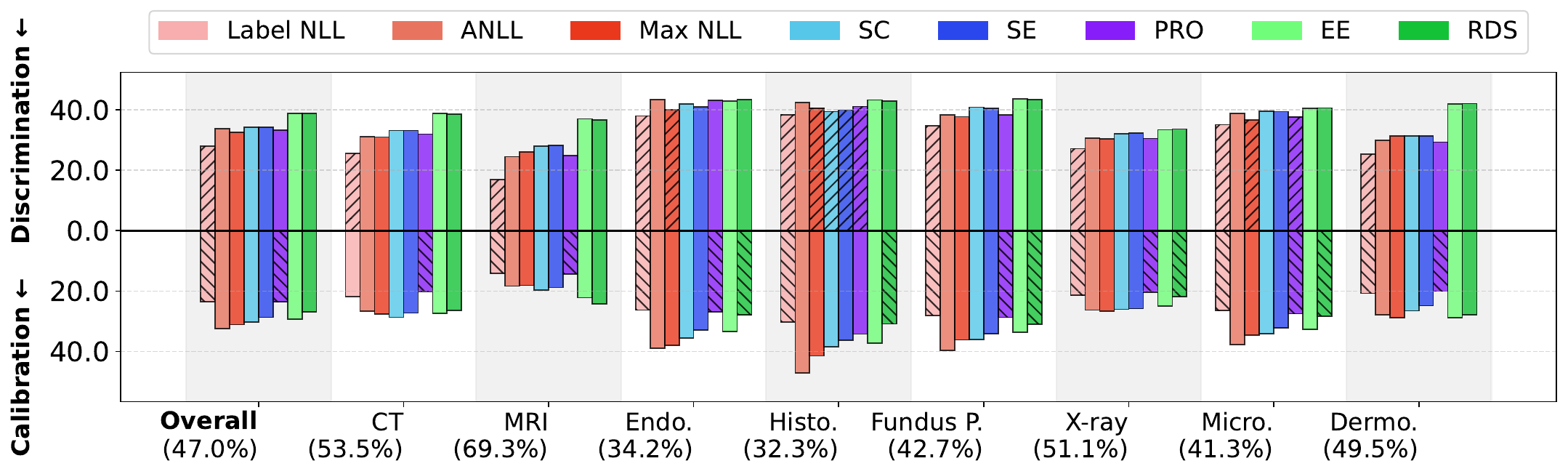}
\caption{Inverse discrimination (1-AUROC, top) and calibration (average ECE and Brier, bottom) for \texttt{Qwen2.5-VL-72B} across modalities. AUROC and ECE values statistically equivalent to the best results are marked with diagonal lines. Model accuracy per modality is shown between parentheses below the modality labels. Modalities (instances balanced, 210 samples for each): \textit{Endoscopy (Endo.)}, \textit{Histopathology (Histo.)}, \textit{Fundus Photography (Fundus P.)}, \textit{Microscopy (Micro.)} and \textit{Dermoscopy (Dermo.)}.}
\label{fig:ece_auroc_per_modality_qwen72b}
\end{figure*}

In the standard MCQA setting, Table~\ref{tab:main_table} presents discrimination and calibration performance across models and UE methods.

\paragraph{Discrimination is dominated by Label NLL.} Label NLL is among the top discrimination methods on 11 of 12 models, often by a large margin (e.g., +4.5 AUROC over the next-best method on \texttt{Qwen2.5-VL-72B}, +5.2 on \texttt{Qwen2.5-VL-32B}). This indicates that the probability assigned to the selected option token carries more information about correctness than aggregated token likelihoods (ANLL, Max NLL) or multi-sample methods, despite requiring only a single forward pass.

\paragraph{Calibration is more heterogeneous, with a logit-embedding split.} No single method leads on calibration: PRO is among the best methods for all \texttt{Qwen} models, RDS for the \texttt{Molmo} family, and ANLL/Max NLL across models of varying size (e.g., \texttt{MedGemma-4B/27B}, \texttt{GPT-4.1-Mini}). More striking is the divergence within methods: Label NLL, the strongest discrimination method, is among the \emph{worst} calibrated on several models (e.g., for the \texttt{MedGemma} family), while embedding- and consistency-based methods rarely lead in discrimination but provide more calibrated confidence estimates. 
The two biomedical models are the worst calibrated overall, and \texttt{GPT-4.1-Mini} has competitive AUROC but poor calibration (ECE $>$ 39 across all methods), suggesting that its post-training may produce systematically overconfident probabilities regardless of the extraction method. Discrimination tracks how well a method ranks predictions; calibration tracks the absolute scale of confidence. The two are not interchangeable, and the best method depends on which property matters downstream.

\paragraph{Size effects are family-dependent.} No family improves monotonically with size on both metrics. AUROC scales cleanly with size in the \texttt{Qwen} family (best-method AUROC rises from 65.7 to 72.0) and improves in the \texttt{Molmo} and \texttt{LLaVA} families, but it slightly decreases for \texttt{MedGemma}. Calibration trends are even less consistent: ECE improves substantially with size for \texttt{MedGemma} and \texttt{Molmo}, but actually \emph{worsens} for \texttt{Qwen} (12.5 to 19.7). Within-family variation is often comparable to across-method variation on the same model, reinforcing that family and method choices interact and neither dominates.

\paragraph{Practical guidance.} 
Small-to-medium models paired with logit-based methods can match or exceed much larger systems: \texttt{Qwen2-VL-7B} surpasses the \texttt{Qwen2-VL-32B} and approaches the \texttt{Qwen2-VL-72B} on ANLL and Max NLL across both ECE and AUROC metrics. Since consistency and embedding approaches require multiple generations, logit-based methods are the most compute-efficient. Method choice should follow the application: prioritize discrimination (e.g., Label NLL) for abstention thresholds, and calibration (e.g., PRO, RDS) when displaying confidence to clinicians.

\subsection{Performance across clinical contexts}
\label{subsec:results_fine_grained}

\input{deltas_all_ue}

The aggregate results in Section~\ref{subsec:results_main} combine performance across all imaging modalities, which may hide important differences in how the model or UE method performs on different clinical contexts, such as CT scans, MRIs, endoscopic videos, or histopathology slides. Prior UE benchmarks have not stratified along this axis, leaving open whether a method that looks well-calibrated overall remains so on the modalities where the model is weakest, precisely the cases where a reliable abstention signal would matter most. We address this by stratifying our most accurate model, \texttt{Qwen2.5-VL-72B}, by imaging modality (Figure~\ref{fig:ece_auroc_per_modality_qwen72b}).

UE quality tracks model accuracy closely across modalities. MRI, where accuracy is highest (69.3\%), shows the strongest discrimination and calibration; Histopathology and Endoscopy, where accuracy collapses to 32-34\%, show the worst (ECE~$\approx 40$, AUROC~$<60$ across all methods, except Label NLL which approaches 60). The ranking by UE quality mirrors the ranking by accuracy almost exactly: MRI~$>$~CT~$\approx$~X-ray~$>$~Dermoscopy~$>$~Fundus~$\approx$~Microscopy~$>$~Endoscopy~$>$~Histopathology.

Two patterns within this trend are worth noting. First, modality-level variation in UE quality often exceeds the variation across UE methods within a modality: the gap between MRI and Histopathology, for any single method, is larger than the gap between the best and worst methods on the overall data. Second, in the lowest-accuracy modalities (Endoscopy, Histopathology with 34.2\%/32.3\%), all methods perform poorly and none extracts a useful signal, indicating that UE methods do not fail uniformly: they fail wherever the model fails.

This relationship is also reflected quantitatively in the correlation between accuracy and UE performance across modalities (Appendix~\ref{appendix:modality_results}). In particular, a substantial subset of model/UE method combinations shows a strong positive correlation ($\rho \geq 0.5$) between accuracy and AUROC. However, this relationship is not consistent across all models. For example, the correlation can be negligible for \texttt{MedGemma-4B} with SE ($\rho = 0.055$) and strongly negative for \texttt{Molmo-7B} with Label NLL ($\rho = -0.495$). The correlation between accuracy and ECE is more mixed, with no similarly consistent pattern. These results therefore suggest that the association between model accuracy and UE quality is strongest for discrimination, but is not universal across models and methods.

This has a direct consequence for deployment: UE is least informative in exactly the contexts where it would be most useful, namely, the contexts where the model struggles. These results therefore do not support treating UE as a safety net for catching model errors in such cases. Whether richer training data, modality-specific encoders, or training-based UE approaches \citep{ICLR2024_ad9d6ab1} can decouple UE quality from accuracy in low-accuracy domains remains an open question for future work.

\section{NOTA Results}
\label{sec:NOTA}

\subsection{Is UE robust to NOTA?}
\label{subsec:nota_main}

We evaluate the hypothesis \emph{H1} introduced in Section \ref{subsec:nota_methodology} by examining how accuracy and calibration jointly shift under the two NOTA perturbations (Table~\ref{tab:deltas_all_ue}; full AUROC, Brier, and average-uncertainty deltas in Appendix Table ~\ref{app_tab:deltas_all_ue}).

\paragraph{Accuracy} Under \texttt{+nota}, where the correct option remains available alongside the NOTA distractor, accuracy drops modestly across all models (by 0.4 to 6.8 points), with the largest declines for \texttt{Qwen2.5-VL-72B} and \texttt{GPT-4.1-Mini}. Under \texttt{-correct+nota}, where the correct option is replaced by NOTA, accuracy collapses (by 12.3 to 32.7 points), confirming that recognizing the absence of a valid answer is substantially harder than resisting a NOTA distractor.

\paragraph{Calibration} \emph{H1} predicts that uncertainty should rise in proportion to these accuracy drops to preserve calibration. The \texttt{+nota} setting largely satisfies this expectation: ECE changes are small and frequently non-significant, with several model and method pairs even improving (e.g., consistency-based methods for the two models with the steepest +nota accuracy drops). The \texttt{-correct+nota} setting tells a different story. Calibration degrades sharply across all models and UE methods. Appendix Table~\ref{app_tab:deltas_all_ue} shows the mechanism: average uncertainty rarely rises with the accuracy collapse, and for some model and method pairs, it actually decreases. \emph{H1} is therefore contradicted in the regime where it matters most: when the model struggles, as when the correct option is unavailable, it commits confidently to a wrong one. 

\subsection{Can UE predict robustness to NOTA?}
\label{subsec:ue_predict_nota}

\input{nota_thresholding_main}
\begin{figure}[t]
    \begin{minipage}{1\columnwidth}
    \centering
    \includegraphics[width=1\linewidth]
    {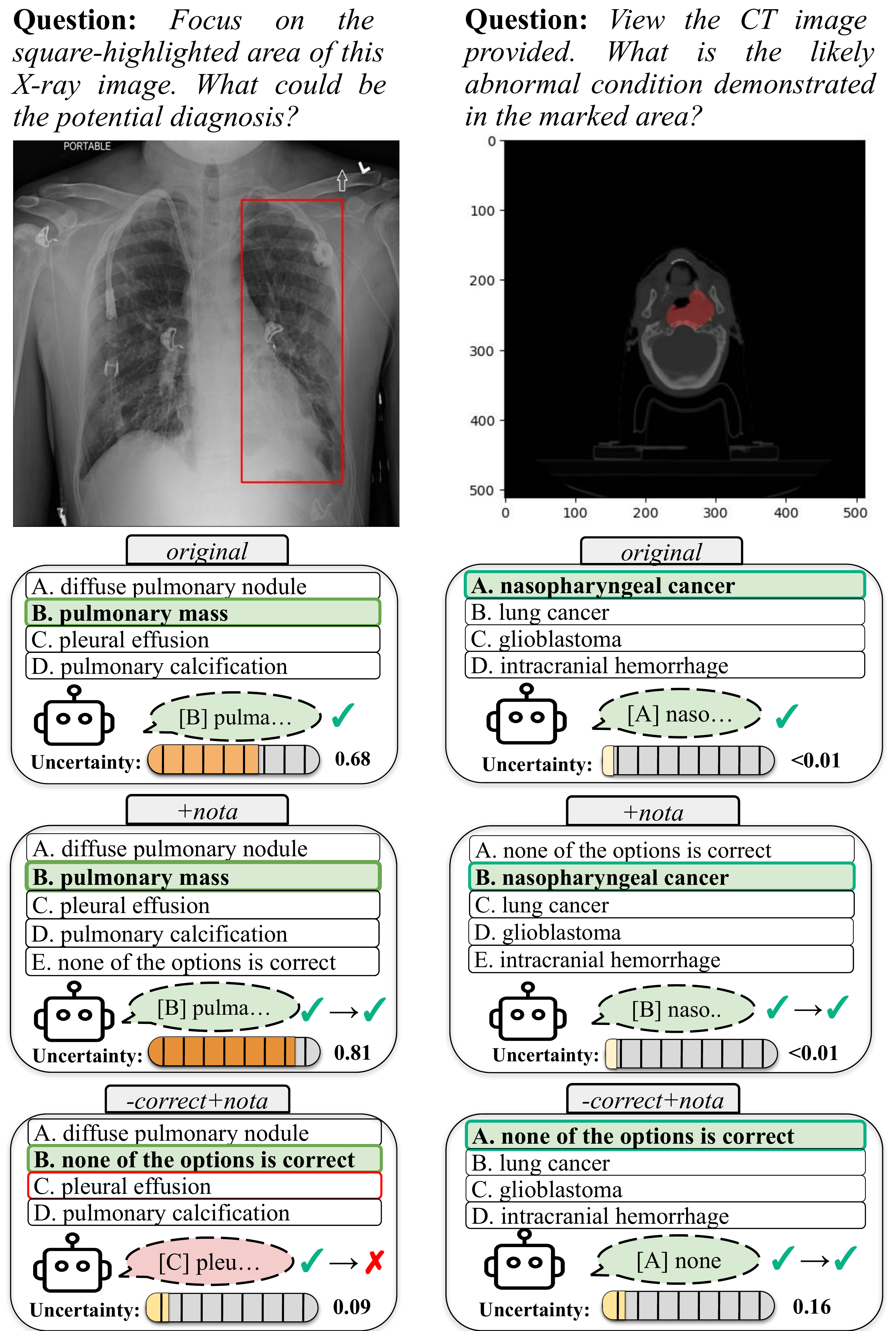}
    \caption{
    Examples of answer shifts and Label NLL uncertainty estimates under \texttt{+nota} and \texttt{-correct+nota} (\texttt{Qwen-2.5-VL-72B} responses). Higher uncertainty on the original input (left) corresponds to a flip under the \texttt{-correct+nota} perturbation, and lower uncertainty to stable predictions.
    }
    \label{fig:nota_example}
    \end{minipage}
\end{figure}

\emph{H1} asked whether uncertainty responds to perturbation. \emph{H2} asks a more demanding question: does uncertainty on the original input predict whether a correct prediction will remain stable or flip under perturbation? This question is relevant not only for deployment, but also for model evaluation. VLMs are typically evaluated on standardized test sets, where inputs follow a fixed format and structure. A model may perform well under this standard format while being sensitive to seemingly minor changes in how the same information is presented. If uncertainty on the original input can predict which correct predictions are vulnerable to such changes, then UE could provide a simple way to augment standard evaluation with a measure of prediction fragility. Here, we test this idea under the NOTA perturbations, as a controlled test of whether UE can identify correct predictions that are fragile to changes in input presentation.

Throughout this analysis, we refer to the uncertainty estimate computed on the original, unperturbed input as \emph{baseline uncertainty}. We first examine average baseline uncertainty for samples whose model answer did or did not flip under the NOTA perturbations. This initial analysis (Appendix~\ref{app_subsubsec:nota_averages}) shows that samples that flip have statistically higher baseline uncertainty than stable samples across all model/UE method combinations under \texttt{+nota} perturbations, and across a large majority of model/UE method combinations under \texttt{-correct+nota} perturbations (85 out of 96 combinations). \texttt{Qwen2.5-VL-72B} shows the largest differences: among samples that were initially correct, those that subsequently flipped to an incorrect answer under \texttt{+nota} had baseline SC uncertainty 4.41 times higher, and baseline SE uncertainty 3.90 times higher, than the average baseline uncertainty across all initially correct samples. These differences provide initial evidence that baseline uncertainty contains information about the subsequent stability of model responses, motivating a more direct evaluation of whether it can predict correct-to-wrong flips.

Following this initial evidence, which is based on average differences between stable and flipped prediction groups, we conduct a direct binary classification experiment. Specifically, we evaluate whether baseline uncertainty can distinguish initially correct predictions that remain stable from those that subsequently flip from correct to incorrect under perturbation. We present AUROC and AUPRC results for the \texttt{+nota} and \texttt{-correct+nota} modifications for \texttt{Qwen2.5-VL-72B} and \texttt{GPT-4.1-Mini} in Table~\ref{tab:nota_thresholds_main_text}. AUROC measures how well uncertainty ranks flipped versus stable predictions across all possible thresholds, while AUPRC summarizes the precision-recall trade-off and is particularly informative when the two outcome classes are imbalanced, as in this task. For both models, all UE methods substantially outperform the majority-vote baseline,\footnote{We also evaluate two additional baselines, a uniform coin flip and a coin flip using the empirical output-class probabilities as the prior. Their performance differs negligibly (by less than 0.01), so we report only the simpler majority-vote baseline.} with logits-based methods generally outperforming consistency-based, hybrid, and embedding-based methods. Predicting correct-to-wrong flips appears more difficult in the \texttt{-correct+nota} setting than in the \texttt{+nota} setting, with performance relative to the baseline lower for both models.

Tables~\ref{app_tab:nota_thresholds_appendix_small} and \ref{app_tab:nota_thresholds_appendix_large} in the Appendix present results for the remaining models. The relative performance of UE method families depends on both the model and the perturbation type. The \texttt{-correct+nota} setting is again more difficult: 12 out of 96 model/UE method combinations perform very close to or below the majority-vote baseline, compared with only 2 combinations for \texttt{+nota}. Embedding-based methods (EE, RDS) are below the baseline for \texttt{LLaVA-v1.6-7B}, \texttt{MedGemma-27B}, and \texttt{LLaVA-v1.6-34B}, whereas logits-based, consistency-based, and hybrid methods perform above the baseline for a substantial majority of model/UE method combinations, with only a small number of exceptions.

Overall, these results show that baseline uncertainty contains a useful signal for predicting the fragility of initially correct predictions. The signal is not yet strong or consistent enough for the resulting classification approach to be used as a reliable predictor of individual prediction flips, particularly for some embedding-based methods and in the more difficult \texttt{-correct+nota} setting. Nevertheless, the fact that simple thresholding of a single UE method can distinguish stable from fragile predictions across many model/UE method combinations provides evidence that prediction fragility is reflected in the uncertainty estimates themselves. Future work could strengthen this signal through post-hoc calibration or processing of UE methods, ensembles, and the development of new UE methods specifically targeting prediction fragility.

\paragraph{Implications and caveats.}
These results suggest that UE captures structural information about model fragility, not only per-prediction confidence. A model/UE method pair that does not satisfy \emph{H1} (uncertainty does not rise after perturbation) can still satisfy \emph{H2} (baseline uncertainty separates fragile from stable predictions), and the diagnostic value of UE is therefore broader than the per-prediction safety-signal interpretation would suggest. At the same time, the classification results show that this signal should not yet be interpreted as a reliable predictor of individual flips: performance varies across UE families, models, and perturbation types, and is weaker in the \texttt{-correct+nota} setting. Thus, the main contribution of this analysis is evidence that baseline UE contains information about prediction fragility, rather than a ready-to-deploy fragility detector.

\section{Conclusion}

We introduce a large-scale benchmark for post-hoc uncertainty estimation (UE) in clinical multiple-choice VQA, evaluating eight UE methods across twelve vision-language models. Our central finding is that UE quality is not an intrinsic property of the UE method but tracks the underlying model's competence: discrimination and calibration co-vary with accuracy across imaging modalities, with degradation concentrated precisely where UE would be most clinically useful. Reliable uncertainty estimation is therefore largely confined to domains where the model already excels, complicating the use of UE as a safety net.

We further identify an asymmetric failure under \textit{None of the above} (NOTA) stress tests: when the correct option is removed, accuracy collapses but uncertainty does not rise enough to match it, leaving models confidently committed to wrong answers in precisely the regime where a safety signal would matter the most. A complementary result reframes the utility of UE. Uncertainty measured on the unperturbed input separates predictions that will remain stable under NOTA from those that will flip, in both directions: high-uncertainty correct answers are the ones most likely to break, and high-uncertainty wrong answers the ones most likely to (coincidentally) self-correct. UE thus carries information about which predictions are structurally fragile. Future UE methods should be evaluated not only on how well their scores align with correctness, but on whether they identify which predictions a model would reverse under modest perturbations, a property current calibration and discrimination metrics capture only partially.

\section{Limitations}

\paragraph{Scope of UE methods.} We evaluate post-hoc UE methods, which are model-agnostic and can be applied to any deployed VLM without retraining. This was a deliberate scope choice: post-hoc methods are the most realistic option for clinical practitioners who do not control model training and must work with vendor-supplied systems. We did not evaluate training-based approaches such as probabilistic adapters \citep{11444867} or probabilistic loss fine-tuning \citep{ICLR2024_ad9d6ab1,NEURIPS2025_af835bd1}, nor mechanistic-interpretability and conformal-prediction methods, all of which require either model access or substantially greater technical scaffolding. Extending the benchmark to these methods is a natural next step. We also note that several post-hoc UE methods produce unbounded scores, requiring min-max normalization for cross-method comparison; while this is standard practice, the absence of a principled alternative is itself a gap in the UE literature. Finally, while we compare trends across model families, some families are represented by only a small number of models, limiting the strength of family-level conclusions.

\paragraph{Scope of the evaluation testbed.} Our benchmark uses English multiple-choice clinical VQA. MCQA provides discrete answer spaces and unambiguous ground truth, which is essential for direct comparison across eight UE methods and twelve models, controlling for the confounders that open-ended evaluation introduces (judge model bias, free-form answer matching). This methodological control comes at a cost: MCQA does not capture open-ended clinical reasoning, multilingual deployment, or naturally occurring ambiguity. Our NOTA stress tests partially relax the closed-world assumption by introducing cases where the correct answer is absent, but they remain a controlled approximation of the underlying clinical scenario. We position this benchmark as a controlled diagnostic for UE behaviour rather than a simulation of clinical practice, and extending it to open-ended VQA with human-judged correctness and to non-English clinical settings is a clear direction for future work. The benchmark itself uses GMAI-MMBench, a single (though broad) dataset of 284 sub-datasets across 38 imaging modalities, and we work with a balanced 1,680-sample subset due to compute constraints; extension to additional dataset families would further test generalization.

\paragraph{Dataset diversity.} Demographic metadata (e.g., race/ethnicity, age, sex/gender, and geographic origin) are not available in GMAI-MMBench, preventing subgroup-specific evaluation. Our findings should not be interpreted as establishing equitable reliability or calibration across patient populations.

\paragraph{Clinical validation.} We do not independently verify that every question in GMAI-MMBench is clinically meaningful and answerable by trained clinicians. The dataset's curators report ethical approval and source quality controls, but expert validation of our specific subset would strengthen conclusions about clinical relevance. We see this as the highest-value direction for follow-up work, together with the open-ended extension.

\paragraph{Additional failure modes.} Several failure modes that may affect both model accuracy and UE quality are not isolated in this study, including: naturally occurring (rather than synthetic) cases where no correct option is listed, low image quality, question-option mismatches, and cross-modal inconsistencies. Some of these can be analyzed within the present framework with additional annotations; others require new datasets. We see these as complementary axes for a future, more comprehensive version of this benchmark.

\section*{Acknowledgments}
AF, AT and IC are funded by the project CaRe-NLP with file number NGF.1607.22.014 of the research programme AiNed Fellowship Grants which is (partly) financed by the Dutch Research Council (NWO) under the grant \href{https://doi.org/10.61686/VYLVB06119}{https://doi.org/10.61686/VYLVB06119}. BP is supported by the ERC Consolidator Grant DIALECT (101043235).

\bibliography{anthology}

\appendix
\input{Dataset_Statistics}
\input{Models}
\input{Prompts}
\input{UE_Methods}
\input{NOTA_deltas_results}

\end{document}

%% file: main_table.tex
\begin{table*}[t]
\centering
\scriptsize
\renewcommand{\arraystretch}{1.2}
\setlength{\tabcolsep}{1.5pt}
\newcolumntype{C}{>{\centering\arraybackslash}p{2.9em}}

\begin{tabular}{r C |
 C C C | C C C | C C |
 C C C | C C C | C C }

 & &
\multicolumn{8}{c|}{\textbf{AUROC} $\uparrow$} &
\multicolumn{8}{c}{\textbf{ECE} $\downarrow$} \\
\cmidrule(lr){3-10}
\cmidrule(lr){11-18}

 & &
\multicolumn{3}{c|}{logits} &
\multicolumn{3}{c|}{consistency/hybrid} &
\multicolumn{2}{c|}{embedding} &
\multicolumn{3}{c|}{logits} &
\multicolumn{3}{c|}{consistency/hybrid} &
\multicolumn{2}{c}{embedding} \\

Model & \textbf{Acc.} &
\makecell{Label\\NLL}
 & ANLL & \makecell{Max\\NLL}
 &
SC & SE & PRO &
EE & RDS &
\makecell{Label\\NLL}
 & ANLL & \makecell{Max\\NLL} &
SC & SE & PRO &
EE & RDS \\
MedGemma-4B & 37.2 & \cellcolor[RGB]{214,232,242} \textbf{57.8} & 53.9 & 53.3 & 50.8 & 50.8 & 53.9 & 51.6 & 51.6 & 55.9 & 46.7 & \cellcolor[RGB]{251,243,224} \textbf{43.9} & 59.4 & 59.2 & 44.9 & 52.2 & 52.1 \\
Qwen2-VL-7B & 42.4 & \cellcolor[RGB]{179,212,232} \textbf{65.6} & 64.6 & 65.2 & 63.9 & 63.7 & \cellcolor[RGB]{179,212,232} \textbf{65.7} & 63.2 & 63.6 & \cellcolor[RGB]{239,197,101} \textbf{12.6} & \cellcolor[RGB]{240,198,104} \textbf{13.5} & \cellcolor[RGB]{240,197,102} \textbf{12.9} & 26.9 & 28.7 & \cellcolor[RGB]{239,196,100} \textbf{12.5} & 26.8 & 26.6 \\
Qwen2.5-VL-7B & 42.2 & \cellcolor[RGB]{188,217,234} \textbf{63.6} & 60.0 & 61.9 & 60.9 & 60.9 & 59.9 & 59.1 & 58.9 & 21.1 & 25.0 & 19.2 & 34.1 & 32.8 & \cellcolor[RGB]{241,203,119} \textbf{16.9} & 24.3 & 23.2 \\
Molmo-7B & 33.9 & 49.1 & 48.6 & 49.0 & \cellcolor[RGB]{227,239,246} \textbf{54.9} & \cellcolor[RGB]{226,238,246} \textbf{55.1} & 48.1 & \cellcolor[RGB]{226,238,246} \textbf{55.1} & \cellcolor[RGB]{227,239,246} \textbf{54.9} & 33.3 & 24.7 & 26.9 & 45.1 & 44.1 & 22.8 & 21.4 & \cellcolor[RGB]{240,200,111} \textbf{15.0} \\
LLaVA-v1.6-7B & 35.5 & \cellcolor[RGB]{195,221,236} \textbf{62.1} & 60.2 & 61.0 & 59.5 & 59.6 & 60.2 & 59.4 & 59.3 & 22.2 & 27.5 & \cellcolor[RGB]{243,209,134} \textbf{20.5} & 32.9 & 30.9 & \cellcolor[RGB]{243,210,136} \textbf{20.9} & 30.9 & 26.7 \\
LLaVA-OV-8B & 42.8 & \cellcolor[RGB]{188,217,234} \textbf{63.6} & 60.4 & 61.4 & 61.7 & 62.0 & 60.4 & 58.4 & 58.3 & 20.3 & 43.7 & 32.4 & 27.9 & 25.0 & \cellcolor[RGB]{240,200,111} \textbf{15.0} & 20.0 & 19.8 \\
\midrule
MedGemma-27B & 39.3 & \cellcolor[RGB]{226,238,246} \textbf{55.1} & 54.2 & 54.6 & \cellcolor[RGB]{222,236,245} \textbf{56.1} & \cellcolor[RGB]{221,236,244} \textbf{56.2} & 53.9 & 53.4 & 53.6 & 53.1 & \cellcolor[RGB]{242,206,127} \textbf{18.7} & 27.3 & 39.0 & 35.3 & \cellcolor[RGB]{242,207,128} \textbf{19.0} & 32.3 & 29.8 \\
Qwen2.5-VL-32B & 44.1 & \cellcolor[RGB]{173,209,230} \textbf{67.0} & 59.3 & 58.8 & 61.5 & 61.8 & 59.8 & 60.0 & 59.5 & 22.6 & 19.9 & 20.3 & 32.8 & 30.3 & \cellcolor[RGB]{241,204,121} \textbf{17.6} & 23.0 & \cellcolor[RGB]{241,203,119} \textbf{16.9} \\
LLaVA-v1.6-34B & 39.0 & \cellcolor[RGB]{173,209,230} \textbf{67.0} & 62.3 & 64.6 & 62.8 & 63.2 & 62.4 & 59.1 & 58.8 & 21.2 & 48.8 & 35.9 & 30.8 & 28.1 & \cellcolor[RGB]{242,206,126} \textbf{18.6} & 24.3 & 23.1 \\
\midrule
Molmo-72B & 35.2 & \cellcolor[RGB]{205,227,239} \textbf{59.8} & 51.3 & 49.7 & 55.7 & 55.8 & 51.3 & 54.9 & 55.0 & 34.6 & \cellcolor[RGB]{239,195,97} \textbf{11.9} & 15.8 & 44.4 & 41.9 & 14.4 & 24.1 & \cellcolor[RGB]{239,194,93} \textbf{10.8} \\
Qwen2.5-VL-72B & 47.0 & \cellcolor[RGB]{150,196,223} \textbf{72.0} & 66.3 & 67.5 & 65.7 & 65.8 & 66.6 & 61.2 & 61.2 & \cellcolor[RGB]{243,209,134} \textbf{20.5} & 31.2 & 30.2 & 28.1 & 26.4 & \cellcolor[RGB]{242,208,131} \textbf{19.7} & 26.1 & 23.1 \\
GPT-4.1-Mini & 43.0 & \cellcolor[RGB]{185,216,233} \textbf{64.2} & 62.8 & 63.3 & 57.1 & 57.2 & 62.9 & 57.0 & 56.8 & 39.8 & 43.2 & \cellcolor[RGB]{250,237,209} \textbf{39.4} & 42.6 & \cellcolor[RGB]{250,237,209} \textbf{39.6} & 41.5 & 41.8 & \cellcolor[RGB]{250,238,210} \textbf{39.7} \\
\end{tabular}
\caption{AUROC (discrimination ↑) and ECE (calibration ↓) results for combinations of VLMs and UE methods. Small models ($\leq$8B) are on top, medium (8B–35B) in the middle, and large models ($>$35B) are at the bottom. Logits-based UE uses one inference run; consistency/embedding-based require multiple. Bold values are statistically equivalent to the best; color density shows AUROC (blue, higher is better) and ECE (orange, lower is better).}
\label{tab:main_table}
\end{table*}

%% file: deltas_all_ue.tex
\begin{table*}[t]
\centering
\scriptsize
\renewcommand{\arraystretch}{1.2}
\setlength{\tabcolsep}{2pt}
\newcolumntype{C}{>{\centering\arraybackslash}p{2.6em}}

\begin{tabular}{r 
 C | C C C C C C C C |
 C | C C C C C C C C }

 &
\multicolumn{9}{c|}{\textbf{\texttt{+nota}}} &
\multicolumn{9}{c}{\textbf{\texttt{-correct+nota}}} \\

Model & Acc. & Label NLL
 & ANLL & Max NLL
 & SC & SE & PRO &
EE & RDS & Acc. &
Label NLL
 & ANLL & Max NLL &
SC & SE & PRO &
EE & RDS \\
\hline
MedGemma-4B & \cellcolor[RGB]{253,250,243} -1.6 & -1.2 & +0.5 & -0.7 & \cellcolor[RGB]{244,248,251} -1.8 & \cellcolor[RGB]{242,247,251} -2.0 & -0.2 & +0.1 & -0.5 & \cellcolor[RGB]{235,178,52} -30.8 & \cellcolor[RGB]{237,187,76} +27.2 & \cellcolor[RGB]{234,177,50} +31.1 & \cellcolor[RGB]{237,188,78} +26.9 & \cellcolor[RGB]{235,181,60} +29.6 & \cellcolor[RGB]{236,185,70} +27.9 & \cellcolor[RGB]{235,178,52} +30.6 & \cellcolor[RGB]{235,180,56} +30.0 & \cellcolor[RGB]{236,183,65} +28.5 \\
Qwen2-VL-7B & \cellcolor[RGB]{253,249,240} -2.4 & +0.6 & -0.1 & -0.1 & +0.2 & +0.9 & +0.6 & \cellcolor[RGB]{253,250,242} +2.0 & +1.7 & \cellcolor[RGB]{237,186,73} -27.5 & \cellcolor[RGB]{242,207,130} +19.0 & \cellcolor[RGB]{243,210,136} +17.9 & \cellcolor[RGB]{243,210,138} +17.6 & \cellcolor[RGB]{242,207,128} +19.1 & \cellcolor[RGB]{250,238,210} +6.8 & \cellcolor[RGB]{250,238,210} +6.8 & \cellcolor[RGB]{248,229,188} +10.1 & \cellcolor[RGB]{249,232,194} +9.1 \\
Qwen2.5-VL-7B & \cellcolor[RGB]{254,251,246} -1.5 & \cellcolor[RGB]{253,249,240} +2.4 & +1.4 & \cellcolor[RGB]{253,250,243} +1.8 & -0.2 & -0.6 & \cellcolor[RGB]{253,250,242} +1.9 & +0.8 & +0.7 & \cellcolor[RGB]{233,173,38} -32.7 & \cellcolor[RGB]{233,172,36} +33.2 & \cellcolor[RGB]{236,182,62} +29.2 & \cellcolor[RGB]{234,174,42} +32.2 & \cellcolor[RGB]{233,173,38} +32.8 & \cellcolor[RGB]{233,172,36} +33.0 & \cellcolor[RGB]{237,188,78} +26.7 & \cellcolor[RGB]{236,184,68} +28.3 & \cellcolor[RGB]{241,203,118} +20.6 \\
Molmo-7B & -0.4 & \cellcolor[RGB]{252,246,232} +3.4 & \cellcolor[RGB]{198,223,237} -8.5 & \cellcolor[RGB]{210,230,241} -6.9 & +1.4 & +1.4 & \cellcolor[RGB]{194,221,236} -9.3 & +0.1 & -0.4 & \cellcolor[RGB]{234,174,42} -32.3 & \cellcolor[RGB]{230,159,0} +38.7 & \cellcolor[RGB]{240,200,110} +21.8 & \cellcolor[RGB]{238,192,89} +24.8 & \cellcolor[RGB]{232,169,28} +34.3 & \cellcolor[RGB]{231,165,18} +35.8 & \cellcolor[RGB]{246,223,170} +12.7 & \cellcolor[RGB]{235,180,58} +29.7 & \cellcolor[RGB]{240,197,102} +23.1 \\
LLaVA-v1.6-7B & \cellcolor[RGB]{252,246,232} -3.4 & -0.6 & +1.5 & -0.6 & \cellcolor[RGB]{224,237,245} -4.6 & \cellcolor[RGB]{234,243,248} -3.1 & +1.4 & 0.0 & -1.1 & \cellcolor[RGB]{242,207,128} -19.3 & \cellcolor[RGB]{242,207,130} +18.8 & \cellcolor[RGB]{241,203,118} +20.8 & \cellcolor[RGB]{242,206,126} +19.6 & \cellcolor[RGB]{242,205,124} +19.9 & \cellcolor[RGB]{242,206,126} +19.4 & \cellcolor[RGB]{242,205,124} +19.9 & \cellcolor[RGB]{240,201,112} +21.7 & \cellcolor[RGB]{242,206,126} +19.6 \\
LLaVA-OV-8B & \cellcolor[RGB]{251,242,222} -4.9 & \cellcolor[RGB]{251,243,224} +4.7 & \cellcolor[RGB]{250,238,210} +6.8 & \cellcolor[RGB]{251,241,220} +5.4 & +0.2 & \cellcolor[RGB]{252,247,234} +3.1 & \cellcolor[RGB]{250,238,211} +6.5 & \cellcolor[RGB]{252,245,230} +3.9 & \cellcolor[RGB]{253,247,236} +3.0 & \cellcolor[RGB]{238,192,89} -25.0 & \cellcolor[RGB]{238,191,86} +25.6 & \cellcolor[RGB]{236,185,70} +27.9 & \cellcolor[RGB]{237,188,78} +26.8 & \cellcolor[RGB]{240,200,110} +21.9 & \cellcolor[RGB]{241,201,113} +21.4 & \cellcolor[RGB]{240,198,105} +22.6 & \cellcolor[RGB]{240,197,102} +23.2 & \cellcolor[RGB]{244,215,150} +16.0 \\
MedGemma-27B & \cellcolor[RGB]{253,250,242} -2.0 & +1.7 & \cellcolor[RGB]{228,240,246} -4.0 & \cellcolor[RGB]{230,241,247} -3.9 & \cellcolor[RGB]{226,238,246} -4.3 & \cellcolor[RGB]{232,242,248} -3.6 & \cellcolor[RGB]{230,241,247} -3.8 & \cellcolor[RGB]{230,241,247} -3.9 & \cellcolor[RGB]{226,238,246} -4.5 & \cellcolor[RGB]{237,186,72} -27.6 & \cellcolor[RGB]{235,181,60} +29.4 & \cellcolor[RGB]{245,216,154} +15.3 & \cellcolor[RGB]{247,226,179} +11.2 & \cellcolor[RGB]{242,207,130} +18.8 & \cellcolor[RGB]{241,204,121} +20.2 & \cellcolor[RGB]{250,237,208} +7.1 & \cellcolor[RGB]{248,230,190} +9.7 & \cellcolor[RGB]{251,242,222} +4.9 \\
Qwen2.5-VL-32B & \cellcolor[RGB]{253,247,236} -2.8 & +0.7 & \cellcolor[RGB]{214,232,242} -6.2 & \cellcolor[RGB]{226,238,246} -4.3 & -1.7 & 0.0 & \cellcolor[RGB]{232,242,248} -3.5 & -1.6 & \cellcolor[RGB]{224,237,245} -4.8 & \cellcolor[RGB]{237,189,80} -26.5 & \cellcolor[RGB]{239,195,96} +24.1 & \cellcolor[RGB]{244,215,150} +16.0 & \cellcolor[RGB]{243,210,138} +17.7 & \cellcolor[RGB]{239,195,97} +23.8 & \cellcolor[RGB]{239,193,92} +24.6 & \cellcolor[RGB]{245,217,156} +15.1 & \cellcolor[RGB]{241,201,113} +21.4 & \cellcolor[RGB]{245,219,160} +14.3 \\
LLaVA-v1.6-34B & \cellcolor[RGB]{253,248,238} -2.5 & -1.2 & \cellcolor[RGB]{253,249,240} +2.2 & -0.6 & \cellcolor[RGB]{238,245,249} -2.6 & -1.3 & +1.4 & \cellcolor[RGB]{253,249,240} +2.2 & +1.8 & \cellcolor[RGB]{238,189,81} -26.0 & \cellcolor[RGB]{237,188,78} +26.9 & \cellcolor[RGB]{236,183,65} +28.5 & \cellcolor[RGB]{236,185,70} +28.1 & \cellcolor[RGB]{238,189,81} +26.1 & \cellcolor[RGB]{237,188,78} +26.7 & \cellcolor[RGB]{237,187,76} +27.1 & \cellcolor[RGB]{236,184,68} +28.4 & \cellcolor[RGB]{240,198,104} +22.8 \\
Molmo-72B & \cellcolor[RGB]{254,253,250} -0.8 & \cellcolor[RGB]{236,244,249} -3.0 & -0.5 & -0.2 & -1.7 & -1.6 & -0.1 & -1.4 & +1.0 & \cellcolor[RGB]{237,187,76} -27.0 & \cellcolor[RGB]{238,190,84} +25.8 & \cellcolor[RGB]{245,217,156} +15.1 & \cellcolor[RGB]{251,241,218} +5.6 & \cellcolor[RGB]{238,191,86} +25.5 & \cellcolor[RGB]{238,192,88} +25.3 & \cellcolor[RGB]{253,247,236} +2.8 & \cellcolor[RGB]{241,202,116} +20.9 & \cellcolor[RGB]{241,204,121} +20.1 \\
Qwen2.5-VL-72B & \cellcolor[RGB]{250,238,210} -6.8 & \cellcolor[RGB]{253,248,238} +2.5 & \cellcolor[RGB]{251,242,222} +5.1 & \cellcolor[RGB]{251,241,220} +5.2 & \cellcolor[RGB]{220,235,244} -5.2 & \cellcolor[RGB]{232,242,248} -3.6 & \cellcolor[RGB]{252,247,234} +3.1 & \cellcolor[RGB]{253,250,243} +1.8 & +1.1 & \cellcolor[RGB]{244,216,152} -15.6 & \cellcolor[RGB]{247,224,174} +12.1 & \cellcolor[RGB]{242,207,130} +18.8 & \cellcolor[RGB]{245,218,158} +14.7 & \cellcolor[RGB]{253,248,238} +2.5 & \cellcolor[RGB]{252,246,232} +3.4 & \cellcolor[RGB]{241,204,121} +20.1 & \cellcolor[RGB]{243,210,138} +17.6 & \cellcolor[RGB]{243,211,140} +17.4 \\
GPT-4.1-Mini & \cellcolor[RGB]{250,238,211} -6.5 & \cellcolor[RGB]{252,245,230} +3.7 & \cellcolor[RGB]{252,245,230} +3.7 & \cellcolor[RGB]{253,250,242} +2.0 & \cellcolor[RGB]{184,215,233} -10.6 & \cellcolor[RGB]{206,227,240} -7.3 & \cellcolor[RGB]{252,246,232} +3.4 & +0.3 & +0.1 & \cellcolor[RGB]{247,224,174} -12.3 & \cellcolor[RGB]{247,225,176} +12.0 & \cellcolor[RGB]{247,225,176} +11.9 & \cellcolor[RGB]{246,223,170} +13.0 & \cellcolor[RGB]{247,226,178} +11.5 & \cellcolor[RGB]{247,226,178} +11.7 & \cellcolor[RGB]{247,225,176} +12.0 & \cellcolor[RGB]{247,225,176} +11.8 & \cellcolor[RGB]{247,225,176} +12.0 \\
\end{tabular}
\caption{Change in Accuracy (↑) and ECE (↓) for the \texttt{+nota} and \texttt{-correct+nota} input modifications per model and per UE method, measured relative to the \texttt{original} benchmark (Table~\ref{tab:main_table}). Statistically significant differences are shaded with blue (orange) indicating higher (lower) performance for each metric.}
\label{tab:deltas_all_ue}
\end{table*}

%% file: nota_thresholding_main.tex
\begin{table}[t]
\centering
\newcolumntype{C}{>{\centering\arraybackslash}p{2.9em}}
\resizebox{0.8\columnwidth}{!}{%
\begin{tabular}{r @{\hskip 0.8cm} CC @{\hskip 0.8cm} CC }
\textbf{UE} & \multicolumn{2}{c}{\textbf{\texttt{+nota}}} & \multicolumn{2}{c}{\textbf{\texttt{-correct+nota}}} \\
\textbf{Method} & AUROC & AUPRC & AUROC & AUPRC \\
\midrule
\multicolumn{4}{l}{Qwen2.5-VL-72B}\\
\midrule
\textit{majority class}  & 0.500 & 0.251 & 0.500 & 0.600  \\
\cmidrule{2-5}
LabelNLL & 0.798 & 0.616 & 0.648 & 0.675  \\
ANLL & 0.769 & 0.554 & 0.604 & 0.664  \\
MaxNLL & 0.799 & 0.572 & 0.614 & 0.683  \\
SC & 0.771 & 0.547 & 0.593 & 0.652  \\
SE & 0.773 & 0.562 & 0.591 & 0.648  \\
PRO & 0.764 & 0.538 & 0.594 & 0.649  \\
EE & 0.713 & 0.445 & 0.577 & 0.672  \\
RDS & 0.717 & 0.450 & 0.575 & 0.666  \\
\midrule
\multicolumn{4}{l}{GPT-4.1-Mini}\\
\midrule
\textit{majority class}  & 0.500 & 0.293 & 0.500 & 0.555  \\
\cmidrule{2-5}
LabelNLL & 0.773 & 0.548 & 0.626 & 0.630  \\
ANLL & 0.750 & 0.513 & 0.613 & 0.627  \\
MaxNLL & 0.754 & 0.524 & 0.613 & 0.620  \\
SC & 0.666 & 0.428 & 0.562 & 0.593  \\
SE & 0.667 & 0.432 & 0.562 & 0.595  \\
PRO & 0.749 & 0.511 & 0.613 & 0.627  \\
EE & 0.669 & 0.466 & 0.532 & 0.593  \\
RDS & 0.669 & 0.465 & 0.529 & 0.594  \\
\bottomrule
\end{tabular}}
\caption{Performance of uncertainty estimation methods in predicting correct-to-wrong answer flips for Qwen2.5-VL-72B (top) and GPT-4.1-Mini (bottom). \textit{majority class} rows show the baseline AUROC and AUPRC per model, accounting for class imbalances.}
\label{tab:nota_thresholds_main_text}
\end{table}

%% file: Dataset_Statistics.tex
\section{Dataset Statistics}
\label{app_sec:dataset_statistics}

We conduct our experiments using the publicly released GMAI-MMBench benchmark \citep{NEURIPS2024_ab7e02fd}, available under the Apache 2.0 license. The benchmark comprises 284 ethically approved datasets, 268 from public sources and 16 contributed by hospitals. The test suite contains eight imaging modalities: CT, MRI, Endoscopy, Histopathology, Fundus Photography, X-ray, Microscopy, and Dermoscopy, with 210 samples per modality. 
Our use is consistent with the benchmark’s research purpose: we use the data only for non-commercial benchmarking and do not redistribute patient data.

%% file: Models.tex
\section{Model Details and Licenses}
\label{app_sec:models}

\begin{table*}[t]
\centering
\scriptsize
\begin{tabular}{lll}
\textbf{Model} & \textbf{Repository} & \textbf{License / Terms} \\
\hline
LLaVA-v1.6-7B & \texttt{llava-hf/llava-v1.6-mistral-7b-hf} & Apache License 2.0 \\
LLaVA-v1.6-34B & \texttt{llava-hf/llava-v1.6-34b-hf} & Not explicitly specified on model card \\
LLaVA-OV-8B & \texttt{lmms-lab/LLaVA-OneVision-1.5-8B-Instruct} & Apache License 2.0 \\
Qwen2-VL-7B & \texttt{Qwen/Qwen2-VL-7B-Instruct} & Apache License 2.0 \\
Qwen2.5-VL-7B & \texttt{Qwen/Qwen2.5-VL-7B-Instruct} & Apache License 2.0 \\
Qwen2.5-VL-32B & \texttt{Qwen/Qwen2.5-VL-32B-Instruct} & Apache License 2.0 \\
Qwen2.5-VL-72B & \texttt{Qwen/Qwen2.5-VL-72B-Instruct} & Qwen License (\href{https://huggingface.co/Qwen/Qwen2.5-VL-72B-Instruct/blob/main/LICENSE}{link}) \\
Molmo-7B & \texttt{allenai/Molmo-7B-D-0924} & Apache License 2.0 \\
Molmo-72B & \texttt{allenai/Molmo-72B-0924} & Apache License 2.0 \\
MedGemma-4B & \texttt{google/medgemma-4b-it} & Health AI Developer Foundations Terms \\
MedGemma-27B & \texttt{google/medgemma-27b-it} & Health AI Developer Foundations Terms \\
GPT-4.1-Mini & - & Proprietary (OpenAI Service Terms) \\
\end{tabular}
\caption{Models, repositories, and licenses.}
\label{tab:model_licenses}
\end{table*}

We access all open-weight models through Hugging Face (\url{https://huggingface.co/}). Table~\ref{tab:model_licenses} lists the repositories and associated licenses. \texttt{GPT-4.1-Mini} is proprietary and accessed via API under OpenAI’s service terms. We perform inference on a cluster equipped with three H100 GPUs and with a batch size of 32 for all models except \texttt{GPT-4.1-Mini}. Small and medium models (3B–32B parameters) complete inference in approximately 1–4 hours, while the largest models require between 20–23 hours. Our use is consistent with the intended research and evaluation use of these models: we perform inference-only benchmarking and do not redistribute weights or restricted outputs.

%% file: Prompts.tex
\section{Prompts and Answer Extraction}
\label{app_sec:prompts}

\paragraph{Model Responses} 
In order to get the model responses to the input question, we employ the following prompt (example of an instantiation in parenthesis):

\input{prompt_inference}

\paragraph{Mapping Responses to the Corresponding Option} 

\input{extraction_details}

We use the following prompt to map model responses that do not follow the required format (i.e., the letter of the selected option in square brackets) to the corresponding answer choice. Placeholders in curly brackets.

\input{prompt_extraction}

Extraction rates (regex-extaction, LLM-extraction, and failures) are shown in Table~\ref{tab:extraction_details}. Failure rates are 7.91\% for \texttt{Qwen2.5-VL-72B}, 2.93\% for \texttt{LLaVA-OV-8B}, and $\leq$0.4\% for all other models. We manually analyse these failures to identify cases where models refuse to answer or produce non-extractable outputs, revealing three main categories: mismatches between questions and answer options, mismatches between questions and images, and questions that are unanswerable due to ambiguity or poor image quality. Some models explicitly refuse to answer, even without a “none” option, and failure types vary across models. \texttt{MedGemma-27B} produces the highest number of explicit refusals, while \texttt{MedGemma-4B} frequently outputs statements such as “I am not an expert.” Model families such as LLaVA-v1.6 and Molmo produce no refusals. Task-specific patterns include repeated responses from \texttt{Qwen2.5-VL-32B} in severity grading and lymphocyte counting, and “Need more context” outputs for lymphocyte counting from multiple models. Aggregating unique refusal cases (excluding gibberish), most occur in counting, severity grading, and histopathology tasks.

%% file: prompt_inference.tex
\begin{quote}
\small
Question: (Examine the image and choose the option that most accurately reflects the depicted severity.) \\
Provide a concise answer and start your response with the letter of the selected option in square brackets. Options: \\
( \\
$[$A$]$ no diabetic retinopathy \\
$[$B$]$ mild (or early) nonproliferative diabetic retinopathy \\
$[$C$]$ advanced proliferative diabetic retinopathy \\
$[$D$]$ very severe nonproliferative diabetic retinopathy \\
) \\
Your Answer:
\end{quote}

%% file: extraction_details.tex
\begin{table}[t]
\centering
\scriptsize
\begin{tabular}{rccc}
\textbf{Model} & \textbf{Regex} & \textbf{LLM} & \textbf{Extraction} \\
 & \textbf{extracted} & \textbf{extracted} & \textbf{failed} \\

\hline
MedGemma-4B & 99.97 & 0.00 & 0.03 \\
Qwen2-VL-7B & 100 & 0.00 & 0.00 \\
Qwen2.5-VL-7B & 99.39 & 0.30 & 0.31 \\
Molmo-7B & 100 & 0.00 & 0.00 \\
LLaVA-v1.6-7B & 100.00 & 0.00 & 0.00 \\
LLaVA-OV-8B & 92.46 & 1.75 & 2.93 \\
MedGemma-27B & 96.53 & 3.12 & 0.35 \\
Qwen2.5-VL-32B & 99.14 & 0.46 & 0.40 \\
LLaVA-v1.6-34B & 100 & 0.00 & 0.00 \\
Molmo-72B & 99.98 & 0.02 & 0.00 \\
Qwen2.5-VL-72B & 87.29 & 4.80 & 7.91 \\
GPT-4.1-Mini & 100 & 0.00 & 0.00 \\
\hline
\end{tabular}
\caption{Ratios (in \%) of inference runs where answers were extracted via regular expressions (“Regex-extracted”), extracted by an LLM (“LLM-extracted”), or failed to extract (“Extraction-failed”). Each sample includes up to 40 inference runs (4 shuffled option orders × 10 stochastic decodes; as explained in Section~\ref{sec:methodology}). The maximum LLM-extracted ratio is 4.8\%. Larger models show fewer instruction-following outputs and more extraction failures (gibberish or refusals).}
\label{tab:extraction_details}
\end{table}

%% file: prompt_extraction.tex
\begin{quote}
\small
You will be given a multiple-choice question with between 2 and 6 answer options (A-F) and a free-form response. \\
Question: \{question\} \\
Options: \\
\{options\} \\
Response: \{response\} \\

Your task is to determine which of the four options the response refers to. Output only the corresponding letter in square brackets, like [X]. Do not include any explanation or additional text. \\
Your annotation must be exactly one of the following: [A], [B], [C], [D], [E], or [] where [] is used if no single valid option is chosen, if multiple options are indicated, or if the response does not clearly map to any option. \\
If the response includes a different letter, you must still map it to one of the valid options above, based on the content of the response. Keep in mind that the response may be poorly formatted or contain irrelevant letters —focus only on identifying the most likely intended option. \\
Your annotation:
\end{quote}

%% file: UE_Methods.tex
\section{UE Methods and Evaluation}
\label{app_sec:ue_methods}

In this section, we provide formal definitions, hyperparameter selection and evaluation details for the uncertainty estimation (UE) methods used in our work.

\subsection{Formal definitions for UE methods}
\label{app_subsec:definitions}

\emph{Semantic Entropy} (SE) \citep{kuhn2023semantic} estimates predictive uncertainty by computing the entropy of the empirical distribution over answer options induced by multiple model generations. Given $N$ independent generations and a discrete set of answer options $\mathcal{Y}$, let $n_y$ denote the number of times option $y \in \mathcal{Y}$ is predicted. The empirical probability of option $y$ is defined as
\begin{equation}
p(y) = \frac{n_y}{N}.
\end{equation}
The semantic entropy is then computed as the Shannon entropy of this distribution:
\begin{equation}
\mathrm{SE} = - \sum_{y \in \mathcal{Y}} p(y) \log p(y).
\end{equation}
Higher values of $H_{\mathrm{SE}}$ indicate greater disagreement among model generations and thus higher predictive uncertainty.

\emph{Probability-Only} (PRO) \citep{Nguyen_Gupta_Le_2026} measures predictive uncertainty via the entropy of the sequence probabilities of sampled output sequences. Unlike Semantic Entropy, PRO does not depend on a predefined set of answer options and is therefore not restricted to the multiple-choice question answering (MCQA) setting.

Given $N$ sampled output sequences $\{g_1, \ldots, g_N\}$ and their corresponding sequence probabilities $P(g_i)$ assigned by the model, we define a normalized probability distribution over the sampled sequences as
\begin{equation}
p(g_i) = \frac{P(g_i)}{\sum_{j=1}^{N} P(g_j)}.
\end{equation}
The PRO uncertainty score is then computed as the Shannon entropy of this distribution:
\begin{equation}
\mathrm{PRO} = - \sum_{i=1}^{N} p(g_i) \log p(g_i).
\end{equation}
Higher values of $H_{\mathrm{PRO}}$ indicate greater dispersion in the model’s probability mass across sampled sequences and thus higher predictive uncertainty.

\emph{EigenScore} \citep{ICLR2024_0d1986a6} estimates uncertainty by computing the trace of the covariance matrix of the hidden representations corresponding to the generated outputs. Formally:
\begin{align}
    \frac{1}{N} \sum_{i=1}^{N} \|x_i - \bar{x}\|_2^{2}
\end{align}
where $x_i$ denotes the hidden representation of the $i$-th generated output, 
\(\bar{x} = \frac{1}{N} \sum_{i=1}^{N} x_i\) is the centroid of the sampled representations, and $N$ is the number of generated outputs.

\emph{EigenEmbed} (EE) \citep{nguyen2026distanceneedradialdispersion} applies the same measure to external embeddings, where now $x_i$ is the sentence embedding of the $i$-th generated output, and $\bar{x}$ is the centroid of these embeddings:

\begin{align}
    \mathrm{EE} = \frac{1}{N} \sum_{i=1}^{N} \|x_i - \bar{x}\|_2^{2}
\end{align}

\emph{Radial Dispersion Score} (RDS) \citep{nguyen2026distanceneedradialdispersion} measures the total radial dispersion of embeddings from their centroid:

\begin{align}
    \mathrm{RDS} = \sum_{i=1}^{N} \|x_i - \bar{x}\|_1
\end{align}

Here, $x_i$ is the sentence embedding of the $i$-th generated output (same as EE), and $\bar{x}$ is the centroid of these embeddings.

\begin{figure*}[t]
  \centering
  \begin{subfigure}[b]{\linewidth}
  \centering
    \includegraphics[width=0.9\textwidth]{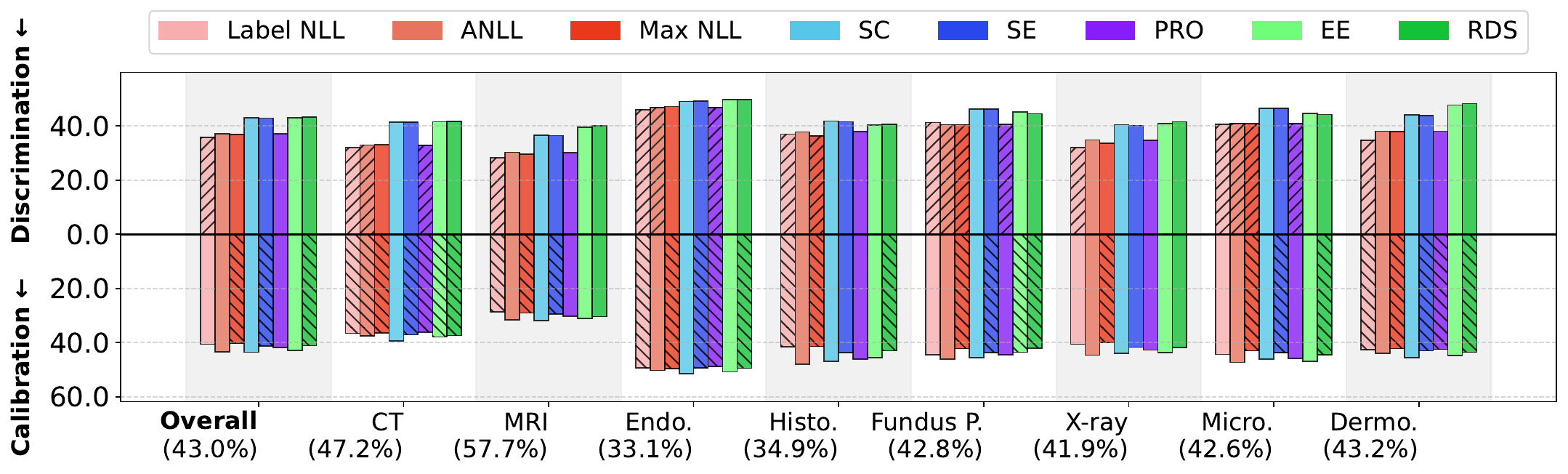}
    \caption{\texttt{GPT-4.1-Mini}}
    \label{fig:ece_auroc_per_modality_gpt}
  \end{subfigure}

  \vspace{10pt} 

  \begin{subfigure}[b]{\linewidth}
  \centering
    \includegraphics[width=0.9\textwidth]{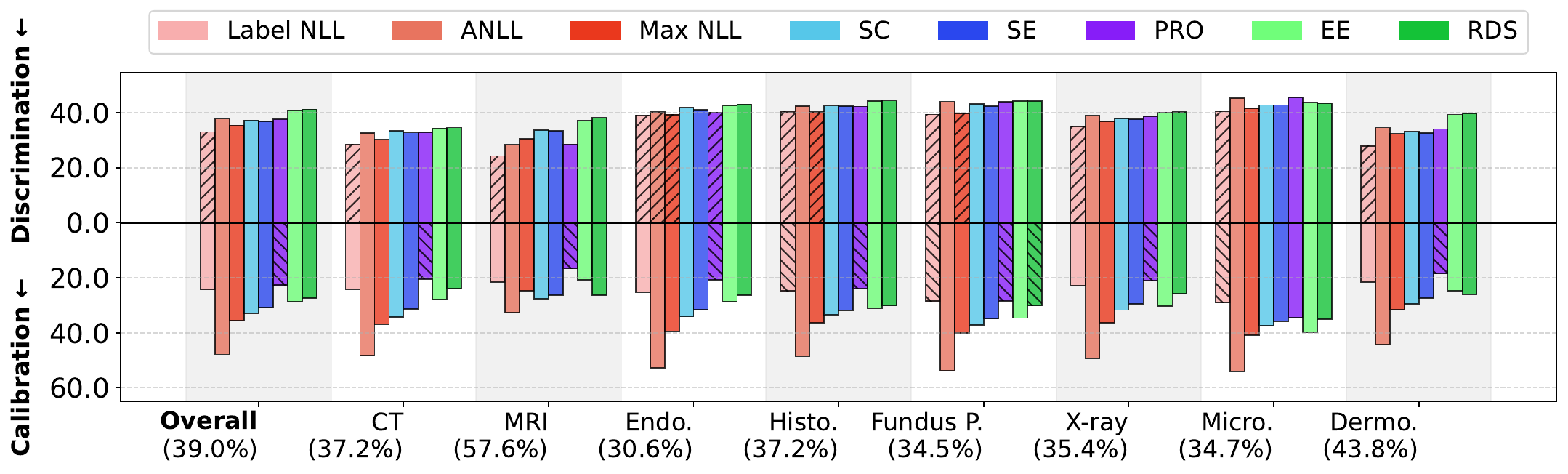}
    \caption{\texttt{LLaVA-v1.6-34B}}
    \label{fig:ece_auroc_per_modality_llava34b}
  \end{subfigure}
  \label{fig:gpt_llava_aroc_ece}
    \caption{Inverse discrimination (1-AUROC, upper half) ↓ and calibration (average ECE and Brier, lower half) ↓. AUROC and ECE values statistically equivalent to the best results are marked with diagonal lines. Model accuracy per modality is shown below the modality labels. Shortened modalities: \textit{Endoscopy (Endo.)}, \textit{Histopathology (Histo.)}, \textit{Fundus Photography (Fundus P.)}, \textit{Microscopy (Micro.)}, and \textit{Dermoscopy (Dermo.)}.}
\end{figure*}

\subsection{Embedding Model Selection}
\label{app_subsec:biomedical_comparison}

\input{main_table_compare_biomedical}

Table~\ref{tab:main_table_compare_biomedical} compares the performance of EE and RDS using a general-purpose\footnote{\href{https://huggingface.co/sentence-transformers/all-MiniLM-L6-v2}{\texttt{all-MiniLM-L6-v2}}} vs. biomedically-trained\footnote{\href{https://huggingface.co/pritamdeka/BioBERT-mnli-snli-scinli-scitail-mednli-stsb}{\texttt{BioBERT-mnli-snli-scinli-scitail-mednli-stsb}}} encoders. The biomedical encoder improves AUROC for most models (except \texttt{Qwen2-VL-7B}), demonstrating the impact of encoder choice on discrimination performance. Calibration (ECE) results vary by model and method. Based on these findings, we use the biomedical encoder for all embedding-based methods in our experiments.
 
\subsection{Normalization of the Uncertainty Estimates}
\label{app_subsec:normalization}
To compute the Expected Calibration Error (ECE) and Brier score, we normalize the output uncertainty estimates to the range [0, 1]. Given that the scale of uncertainty estimates varies across uncertainty estimation (UE) methods and even models (for the same UE method), we adopt a simplified post-hoc calibration approach.

In our experiments, all UE methods yield a minimum value of 0.0. To determine the upper bound, we use a separate partition of 1,000 samples from the GMAI-MMBench test set \cite{NEURIPS2024_ab7e02fd} (evaluations use the validation set; see Subsection~\ref{subsec:dataset}). We compute the maximum uncertainty value for each model-UE method combination and set the upper limit at the 99th percentile to mitigate outlier skew. During experiments, we apply min-max normalization to scale uncertainty estimates between 0 and this upper limit for each pair.

%% file: main_table_compare_biomedical.tex
\begin{table}[t]
\centering
\scriptsize
\renewcommand{\arraystretch}{1.2}
\setlength{\tabcolsep}{2pt}
\newcolumntype{C}{>{\centering\arraybackslash}p{2.4em}}

\begin{tabular}{r 
 C C | C C | C C | C C }

 &
\multicolumn{4}{c|}{\textbf{AUROC} $\uparrow$} &
\multicolumn{4}{c}{\textbf{ECE} $\downarrow$} \\

Model &
EE & EE-bio &  RDS & RDS-bio &
EE & EE-bio &  RDS & RDS-bio \\
\hline
MedGemma-4B  &  \textbf{50.2} & \textbf{51.6} & \textbf{50.1} & \textbf{51.6} & \textbf{43.9} & 53.7 & \textbf{37.9} & 51.3 \\
Qwen2-VL-7B  &  61.8 & \textbf{63.1} & 63.0 & \textbf{63.4} & \textbf{43.5} & 47.0 & \textbf{30.5} & 34.2 \\
Qwen2.5-VL-7B  &  49.3 & \textbf{59.1} & 49.5 & \textbf{58.9} & \textbf{25.2} & 29.3 & \textbf{22.3} & \textbf{21.4} \\
Molmo-7B  &  50.2 & \textbf{55.1} & 50.2 & \textbf{54.9} & \textbf{24.7} & \textbf{25.1} & 18.0 & \textbf{14.3} \\
LLaVA-v1.6-7B  &  50.1 & \textbf{59.4} & 50.0 & \textbf{59.3} & 37.0 & \textbf{34.8} & 31.3 & \textbf{25.8} \\
LLaVA-OV-8B  &  50.8 & \textbf{58.3} & 50.7 & \textbf{58.2} & \textbf{22.5} & 24.1 & \textbf{18.7} & \textbf{17.5} \\
MedGemma-27B  &  52.9 & \textbf{53.4} & \textbf{53.4} & \textbf{53.6} & 32.4 & \textbf{30.9} & 29.6 & \textbf{28.2} \\
Qwen2.5-VL-32B  &  49.2 & \textbf{60.0} & 49.5 & \textbf{59.5} & \textbf{23.4} & 30.0 & \textbf{17.3} & \textbf{15.6} \\
LLaVA-v1.6-34B  &  49.6 & \textbf{59.1} & 49.6 & \textbf{58.8} & \textbf{29.5} & \textbf{28.1} & 27.8 & \textbf{20.9} \\
Molmo-72B  &  49.5 & \textbf{54.9} & 49.4 & \textbf{55.0} & \textbf{26.4} & 28.6 & 14.9 & \textbf{11.7} \\
Qwen2.5-VL-72B  &  46.9 & \textbf{60.3} & 46.8 & \textbf{60.2} & \textbf{29.6} & \textbf{30.9} & 26.8 & \textbf{22.2} \\
GPT-4.1-Mini  &  50.4 & \textbf{57.0} & 50.5 & \textbf{56.8} & \textbf{45.2} & \textbf{44.9} & 42.8 & \textbf{40.0} \\
\end{tabular}
\caption{Comparison of embedding-based methods using the general-purpose vs. biomedically-trained encoder. For each model and UE method combination, AUROC and ECE values statistically equivalent to the best results are \textbf{highlighted in bold}.}
\label{tab:main_table_compare_biomedical}
\end{table}

%% file: NOTA_deltas_results.tex
\section{Additional Results}
\label{app_sec:NOTA}

\subsection{Performance per imaging modality}
\label{appendix:modality_results}
\input{modality_correlations}

Figure~\ref{fig:ece_auroc_per_modality_gpt} shows \texttt{GPT-4.1-Mini}’s UE performance and accuracy per imaging modality. UE performance closely tracks model accuracy, with less variation across UE methods than across modalities. The two highest-accuracy modalities (MRI and CT, with 57.7\% and 47.2\% accuracy respectively) also exhibit the best UE performance, while the lowest-accuracy modality (Endoscopy) has the worst UE performance. A similar trend appears for \texttt{LLaVA-v1.6-34B} (Figure~\ref{fig:ece_auroc_per_modality_llava34b}): MRI (57.6\% accuracy) leads in both calibration and discrimination, while Dermoscopy (43.8\%) and CT (37.2\%) show better calibration and discrimination than lower-accuracy modalities. These results confirm Section~\ref{subsec:results_fine_grained}: UE is least informative precisely where it is most critical, when model performance drops.

\paragraph{Correlation between accuracy and UE performance across modalities}
To further support the analysis in Section~\ref{subsec:results_fine_grained}, Table~\ref{app_tab:modalities_correlation} reports Pearson correlations between accuracy and UE performance (AUROC, ECE) across imaging modalities. Accuracy and AUROC show strong, significant positive correlations ($\rho \geq 0.7$) in 43 of 96 model/UE method combinations, while 74 of 96 exhibit at least moderate positive correlations ($\rho \geq 0.5$), although not all are statistically significant. These results should be interpreted cautiously, as each correlation is computed from only 8 samples (one per modality), and no multiple-testing correction is applied. The strength and reliability of the correlations also appear model-family dependent: MedGemma and Molmo show notably weaker correlations, with the direction even varying across UE methods.

Correlations between accuracy and ECE are weaker and more variable in direction. Strong, significant negative correlations ($\rho \leq -0.7$) occur in 36 of 96 combinations, while 52 of 96 show at least moderate negative correlations ($\rho \leq -0.5$). Thus, although the accuracy--ECE relationship is less consistent than the accuracy--AUROC relationship, the overall negative association remains evident.

\subsection{Is UE robust to NOTA?}
\label{app_subsec:5_1}

\input{deltas_all_UE_appendix}

In subsection~\ref{subsec:nota_main}, we show that ECE drops in the \texttt{-correct+nota} setting because estimated uncertainty fails to increase proportionally with the large accuracy drop (12.3\%–32.7\%; Table~\ref{tab:deltas_all_ue}). This reveals a critical limitation: UE fails to flag the very contexts where it would be most valuable, where the model struggles most. Conversely, in \texttt{+nota}, calibration improves for some cases, particularly with consistency-based methods, as the larger answer space naturally increases uncertainty.

Table~\ref{app_tab:deltas_all_ue} provides the changes in accuracy and average estimated uncertainty for each UE method. AUROC (discrimination) and Brier (calibration) deltas follow the same pattern as ECE:  In \texttt{+nota}, performance improves for many model-UE method combinations, especially consistency-based ones. In \texttt{correct+nota}, however, both AUROC and Brier drop sharply for most combinations, with few exceptions (e.g., PRO improves for 4 of 12 models). These drops occur because accuracy falls dramatically while estimated uncertainty increases only for some model-UE method pairs, and even then, not enough to offset the accuracy loss. This undermines UE reliability in practice (Section~\ref{subsec:nota_main}).

\subsection{Can UE predict robustness to NOTA?}
\label{app_subsec:5_2}

\subsubsection{Comparison between baseline uncertainty estimates}
\label{app_subsubsec:nota_averages}

\input{nota_averages_multiple_models_appendix_small}
\input{nota_averages_multiple_models_appendix_large}

For each model, we compute baseline uncertainty (on the unperturbed input) and partition samples by their behaviour under perturbation: \CtoC{} (initially correct, remains correct, or selects NOTA when the correct option is removed); \CtoW{} (initially correct, flips to a wrong option); \WtoC{} (initially wrong, becomes correct); \WtoW{} (initially wrong, stays wrong). Tables~\ref{tab:5_2_small_models} and \ref{tab:5_2_large_models} report the relative difference between each subset's average baseline uncertainty and the corresponding initially-correct or initially-wrong pool. The results show that baseline uncertainty on unperturbed inputs differs systematically between stable and flipped predictions under NOTA perturbations. 

Consistent with \emph{H2}, samples that flip (\CtoW, \WtoC) show higher baseline uncertainty than stable samples (\CtoC, \WtoW) across all models and both NOTA settings. The effect is strongest under \texttt{+nota}, where the correct option remains available and a flip therefore indicates brittleness: on \texttt{Qwen2.5-VL-72B}, \CtoW{} samples have a relative difference of +4.41 in baseline SC uncertainty and +3.90 in SE compared to the average correct sample. Symmetrically, \WtoC{} samples carry above-average uncertainty among initially-wrong predictions, meaning baseline uncertainty predicts instability in both directions, not only deterioration. Under \texttt{-correct+nota}, where every model loses substantial accuracy, the separation is smaller in magnitude but consistent in sign.

Consistency-based methods (SC, SE) show the clearest separation across all models except MedGemma-4B (Table~\ref{tab:5_2_small_models}), with Label NLL close behind. ANLL, Max NLL, and PRO behave inconsistently: strong on \texttt{Qwen2.5-VL-72B} and \texttt{LLaVA-v1.6-34B}, but near-zero on \texttt{Molmo-72B} (e.g., PRO shows essentially no separation for both \CtoC{} and \CtoW{}). Embedding-based methods (EE, RDS) show the most striking failure: on \texttt{GPT-4.1-Mini}, both are flat across all subsets, despite their non-trivial overall AUROC on this model (Table~\ref{tab:main_table}). Discriminating correctness and predicting fragility are therefore not interchangeable: the best method for one is not necessarily the best for the other.

Two caveats temper this picture. First, the \WtoC{} subset under \texttt{-correct+nota} is small (n=25-139) because models rarely select NOTA, and conclusions about that quadrant should be treated as exploratory. Second, the analysis here uses average differences across subsets; and can be taken as additional analysis to the results in Subsection~\ref{subsec:ue_predict_nota}.

\subsubsection{Baseline uncertainty predicts answer flipping under NOTA}
\label{app_subsubsec:nota_thresholds}

\input{nota_thresholding_multiple_models_appendix_small}
\input{nota_thresholding_multiple_models_appendix_large}

Following the analysis in Section~\ref{subsec:ue_predict_nota}, we report the same binary classification results for the remaining models in Tables~\ref{app_tab:nota_thresholds_appendix_small} and \ref{app_tab:nota_thresholds_appendix_large}. The tables report AUROC and AUPRC for predicting whether an initially correct answer remains stable or flips to an incorrect answer under the \texttt{+nota} and \texttt{-correct+nota} perturbations, using baseline uncertainty computed on the original, unperturbed input. These results provide the complete set of model/UE method combinations and support the model- and perturbation-dependent patterns discussed in Section~\ref{subsec:ue_predict_nota}.

\section{AI Assistants}
We used OpenAI’s ChatGPT to help with minor stylistic revisions and coding tasks. All outputs generated by the model were carefully reviewed and verified by the authors.

%% file: modality_correlations.tex
\begin{table*}[t]
\centering
\scriptsize
\renewcommand{\arraystretch}{1.2}
\setlength{\tabcolsep}{2pt}
\newcolumntype{C}{>{\centering\arraybackslash}p{2.9em}}

\begin{tabular}{r 
 C C C C C C C C |
 C C C C C C C C }

 &
\multicolumn{8}{c|}{\textbf{Acc vs. AUROC}} &
\multicolumn{8}{c}{\textbf{Acc vs. ECE}} \\

Model & Label NLL
 & ANLL & Max NLL
 &
SC & SE & PRO &
EE & RDS &
Label NLL
 & ANLL & Max NLL &
SC & SE & PRO &
EE & RDS \\
\toprule
MedGemma-4B & +52.6 & +55.8 & +55.1 & +5.8 & +5.5 & +56.3 & +52.8 & +56.1 & \cellcolor[RGB]{252,245,230} -74.9 & -27.6 & -25.0 & \cellcolor[RGB]{249,233,198} -81.0 & \cellcolor[RGB]{249,235,202} -80.5 & -27.5 & -50.4 & -47.1 \\
Qwen2-VL-7B & \cellcolor[RGB]{179,213,232} +84.6 & \cellcolor[RGB]{195,222,237} +81.7 & \cellcolor[RGB]{186,216,234} +83.5 & \cellcolor[RGB]{162,203,226} +88.0 & \cellcolor[RGB]{166,205,228} +87.5 & \cellcolor[RGB]{178,212,231} +85.0 & \cellcolor[RGB]{168,206,228} +86.8 & \cellcolor[RGB]{163,204,227} +87.9 & -56.4 & -47.2 & -54.1 & \cellcolor[RGB]{248,228,184} -83.7 & -63.7 & \cellcolor[RGB]{248,251,252} +71.5 & -62.2 & -59.8 \\
Qwen2.5-VL-7B & \cellcolor[RGB]{210,230,241} +78.6 & \cellcolor[RGB]{188,217,234} +83.1 & \cellcolor[RGB]{186,216,234} +83.5 & \cellcolor[RGB]{202,225,238} +80.4 & \cellcolor[RGB]{220,235,244} +77.0 & \cellcolor[RGB]{190,219,235} +82.7 & \cellcolor[RGB]{226,238,246} +75.5 & \cellcolor[RGB]{228,240,246} +75.4 & \cellcolor[RGB]{250,236,206} -79.5 & -70.7 & \cellcolor[RGB]{251,243,224} -76.2 & \cellcolor[RGB]{246,223,172} -86.3 & \cellcolor[RGB]{247,225,176} -85.5 & -65.4 & -42.9 & +6.5 \\
Molmo-7B & -49.5 & +7.3 & -13.3 & +42.7 & +41.9 & -12.6 & +41.6 & +37.2 & +47.8 & +5.1 & +27.0 & -38.1 & -38.3 & +19.2 & -44.7 & -24.0 \\
LLaVA-v1.6-7B & \cellcolor[RGB]{234,243,248} +74.0 & \cellcolor[RGB]{194,221,236} +81.9 & +70.2 & +67.1 & +67.3 & \cellcolor[RGB]{200,224,238} +80.6 & +57.4 & +54.7 & +8.3 & -21.8 & -3.6 & -1.3 & +10.2 & -31.2 & -7.9 & +9.8 \\
LLaVA-OV-8B & \cellcolor[RGB]{220,235,244} +77.0 & +63.6 & \cellcolor[RGB]{208,229,240} +79.2 & \cellcolor[RGB]{200,224,238} +80.9 & \cellcolor[RGB]{208,229,240} +79.1 & +55.4 & +60.8 & +59.0 & \cellcolor[RGB]{249,232,195} -81.6 & \cellcolor[RGB]{244,215,150} -90.4 & \cellcolor[RGB]{244,214,147} -91.0 & \cellcolor[RGB]{246,220,163} -87.8 & \cellcolor[RGB]{249,232,195} -81.6 & -56.0 & -69.8 & +22.9 \\
MedGemma-27B & -5.5 & +26.9 & +41.5 & \cellcolor[RGB]{182,214,232} +84.4 & \cellcolor[RGB]{202,225,238} +80.5 & +22.7 & +27.8 & +28.7 & \cellcolor[RGB]{247,224,174} -85.7 & -18.0 & -41.1 & \cellcolor[RGB]{251,240,216} -77.6 & -68.3 & +24.0 & -26.0 & -7.1 \\
Qwen2.5-VL-32B & \cellcolor[RGB]{195,222,237} +81.5 & +62.1 & +64.3 & \cellcolor[RGB]{208,229,240} +79.2 & \cellcolor[RGB]{206,227,240} +79.6 & +63.0 & \cellcolor[RGB]{238,245,249} +73.2 & \cellcolor[RGB]{234,243,248} +74.3 & \cellcolor[RGB]{244,213,144} -91.8 & -69.3 & \cellcolor[RGB]{253,249,240} -73.0 & \cellcolor[RGB]{242,207,128} -95.0 & \cellcolor[RGB]{241,201,113} -97.6 & \cellcolor[RGB]{254,253,250} -71.1 & \cellcolor[RGB]{246,221,166} -87.6 & -62.9 \\
LLaVA-v1.6-34B & \cellcolor[RGB]{208,229,240} +79.3 & \cellcolor[RGB]{244,248,251} +72.1 & \cellcolor[RGB]{242,247,251} +72.4 & +66.9 & +65.3 & +70.6 & +66.3 & +64.3 & -22.9 & \cellcolor[RGB]{251,241,220} -77.0 & \cellcolor[RGB]{253,250,242} -72.4 & -38.9 & -34.7 & -29.7 & -47.7 & -18.2 \\
Molmo-72B & +57.6 & -18.1 & -32.1 & +45.9 & +44.4 & -15.6 & +68.9 & +67.1 & -20.1 & +23.2 & +57.6 & -35.2 & -33.4 & +63.1 & \cellcolor[RGB]{247,227,182} -84.4 & -36.0 \\
Qwen2.5-VL-72B & \cellcolor[RGB]{154,199,224} +89.6 & \cellcolor[RGB]{166,205,228} +87.4 & \cellcolor[RGB]{168,206,228} +86.8 & \cellcolor[RGB]{216,233,243} +77.8 & \cellcolor[RGB]{214,232,242} +78.1 & \cellcolor[RGB]{172,209,229} +86.2 & +51.5 & +55.4 & \cellcolor[RGB]{249,232,195} -81.3 & \cellcolor[RGB]{249,232,194} -82.1 & \cellcolor[RGB]{245,216,154} -89.6 & \cellcolor[RGB]{250,236,206} -79.7 & \cellcolor[RGB]{250,236,206} -79.6 & \cellcolor[RGB]{250,238,211} -78.3 & \cellcolor[RGB]{250,235,204} -80.2 & -28.2 \\
GPT-4.1-Mini & \cellcolor[RGB]{224,237,245} +76.1 & \cellcolor[RGB]{204,226,239} +79.8 & \cellcolor[RGB]{226,238,246} +75.5 & \cellcolor[RGB]{250,252,253} +71.1 & +69.8 & \cellcolor[RGB]{200,224,238} +80.6 & +51.6 & +48.8 & \cellcolor[RGB]{245,218,158} -89.0 & \cellcolor[RGB]{242,207,130} -94.5 & \cellcolor[RGB]{243,211,140} -92.4 & \cellcolor[RGB]{242,205,124} -95.7 & \cellcolor[RGB]{242,207,130} -94.6 & \cellcolor[RGB]{242,207,130} -94.3 & \cellcolor[RGB]{243,211,140} -92.4 & \cellcolor[RGB]{243,211,140} -92.4 \\
\bottomrule
\end{tabular}
\caption{Pearson correlation ($\rho \times 100$) between Accuracy and AUROC (↑; left) and ECE (↓; right). Blue/orange shading indicates significantly positive/negative correlations ($p \leq 0.05$), with shading intensity proportional to the magnitude of the correlation coefficient ($|\rho|$). Since lower ECE scores indicate better calibration, the expected relationship is positive for AUROC and negative for ECE.}
\label{app_tab:modalities_correlation}
\end{table*}

%% file: deltas_all_UE_appendix.tex
\begin{table*}[t]
\centering
\scriptsize
\renewcommand{\arraystretch}{1.2}
\setlength{\tabcolsep}{2pt}
\newcolumntype{C}{>{\centering\arraybackslash}p{2.9em}}

\begin{tabular}{r 
 C C C C C C C C |
 C C C C C C C C }

 &
\multicolumn{8}{c|}{\textbf{\texttt{+nota}}} &
\multicolumn{8}{c}{\textbf{\texttt{-correct+nota}}} \\

Model & Label NLL
 & ANLL & Max NLL
 &
SC & SE & PRO &
EE & RDS &
Label NLL
 & ANLL & Max NLL &
SC & SE & PRO &
EE & RDS \\
\toprule
MedGemma-4B & -0.8 & +1.4 & +1.9 & \cellcolor[RGB]{238,245,249} +2.6 & \cellcolor[RGB]{238,245,249} +2.6 & +1.4 & +1.9 & +1.8 & \cellcolor[RGB]{244,214,147} -17.2 & \cellcolor[RGB]{250,235,204} -8.2 & \cellcolor[RGB]{247,226,179} -12.2 & \cellcolor[RGB]{252,247,234} -3.4 & \cellcolor[RGB]{252,247,234} -3.4 & \cellcolor[RGB]{250,236,206} -8.1 & \cellcolor[RGB]{249,232,195} -9.4 & \cellcolor[RGB]{251,241,220} -5.8 \\
Qwen2-VL-7B & +0.2 & +0.4 & +0.3 & +0.5 & +0.3 & -0.0 & -1.6 & -1.5 & \cellcolor[RGB]{241,203,118} -22.0 & \cellcolor[RGB]{243,210,138} -18.7 & \cellcolor[RGB]{240,201,112} -23.0 & \cellcolor[RGB]{243,210,138} -18.7 & \cellcolor[RGB]{244,213,144} -18.0 & \cellcolor[RGB]{243,212,142} -18.1 & \cellcolor[RGB]{239,194,94} -25.9 & \cellcolor[RGB]{238,191,86} -27.2 \\
Qwen2.5-VL-7B & +0.1 & -0.1 & +0.4 & +0.7 & +0.9 & -0.1 & -0.5 & -0.7 & \cellcolor[RGB]{245,219,160} -15.2 & \cellcolor[RGB]{251,241,218} -6.0 & \cellcolor[RGB]{246,220,163} -14.6 & \cellcolor[RGB]{247,227,182} -11.8 & \cellcolor[RGB]{247,226,179} -12.2 & \cellcolor[RGB]{251,242,222} -5.3 & \cellcolor[RGB]{240,199,108} -23.6 & \cellcolor[RGB]{242,205,124} -21.1 \\
Molmo-7B & \cellcolor[RGB]{208,229,240} +7.6 & \cellcolor[RGB]{236,244,249} +3.1 & \cellcolor[RGB]{240,246,250} +2.5 & -0.3 & -0.3 & \cellcolor[RGB]{234,243,248} +3.5 & -0.4 & -0.7 & \cellcolor[RGB]{235,181,60} -31.4 & \cellcolor[RGB]{248,229,186} -11.1 & \cellcolor[RGB]{245,219,162} -14.8 & \cellcolor[RGB]{234,177,50} -32.9 & \cellcolor[RGB]{234,174,42} -34.3 & \cellcolor[RGB]{249,233,198} -9.2 & \cellcolor[RGB]{230,159,0} -41.2 & \cellcolor[RGB]{231,164,14} -38.8 \\
LLaVA-v1.6-7B & -0.0 & +2.0 & +1.8 & \cellcolor[RGB]{218,234,243} +5.9 & \cellcolor[RGB]{222,236,245} +5.2 & +1.7 & \cellcolor[RGB]{240,246,250} +2.3 & \cellcolor[RGB]{242,247,251} +2.3 & \cellcolor[RGB]{245,217,156} -15.9 & \cellcolor[RGB]{240,200,110} -23.2 & \cellcolor[RGB]{241,202,116} -22.2 & \cellcolor[RGB]{247,227,182} -11.9 & \cellcolor[RGB]{247,226,178} -12.2 & \cellcolor[RGB]{241,202,116} -22.4 & \cellcolor[RGB]{236,183,64} -30.6 & \cellcolor[RGB]{237,187,76} -28.7 \\
LLaVA-OV-8B & +0.4 & \cellcolor[RGB]{253,250,242} -2.2 & -0.1 & \cellcolor[RGB]{234,243,248} +3.5 & \cellcolor[RGB]{242,247,251} +2.0 & \cellcolor[RGB]{253,249,240} -2.4 & +0.5 & +0.3 & \cellcolor[RGB]{242,207,130} -20.3 & \cellcolor[RGB]{252,244,227} -4.4 & \cellcolor[RGB]{244,214,147} -17.4 & \cellcolor[RGB]{245,216,154} -16.2 & \cellcolor[RGB]{245,216,154} -16.3 & -1.3 & \cellcolor[RGB]{242,207,130} -20.0 & \cellcolor[RGB]{244,213,146} -17.5 \\
MedGemma-27B & -0.7 & +1.7 & +1.4 & \cellcolor[RGB]{238,245,249} +2.9 & \cellcolor[RGB]{238,245,249} +2.7 & +1.7 & +0.2 & +0.2 & \cellcolor[RGB]{252,244,227} -4.5 & \cellcolor[RGB]{243,210,136} -19.2 & \cellcolor[RGB]{237,186,72} -29.3 & \cellcolor[RGB]{242,208,131} -19.8 & \cellcolor[RGB]{243,210,136} -19.0 & \cellcolor[RGB]{245,217,156} -16.0 & \cellcolor[RGB]{250,238,210} -7.2 & \cellcolor[RGB]{248,230,190} -10.6 \\
Qwen2.5-VL-32B & +0.4 & -0.6 & -0.7 & \cellcolor[RGB]{228,240,246} +4.2 & \cellcolor[RGB]{232,242,248} +3.8 & -0.5 & -0.1 & +0.0 & \cellcolor[RGB]{243,210,138} -18.8 & \cellcolor[RGB]{235,180,58} -31.6 & \cellcolor[RGB]{231,166,20} -37.8 & \cellcolor[RGB]{247,226,179} -12.1 & \cellcolor[RGB]{247,226,178} -12.4 & \cellcolor[RGB]{234,175,44} -34.0 & \cellcolor[RGB]{245,218,158} -15.6 & \cellcolor[RGB]{244,213,146} -17.6 \\
LLaVA-v1.6-34B & -1.5 & -1.6 & -0.9 & \cellcolor[RGB]{238,245,249} +2.8 & \cellcolor[RGB]{242,247,251} +2.1 & -1.7 & +0.9 & +0.7 & \cellcolor[RGB]{244,216,152} -16.5 & \cellcolor[RGB]{250,236,206} -7.8 & \cellcolor[RGB]{246,221,166} -14.5 & \cellcolor[RGB]{247,224,174} -13.0 & \cellcolor[RGB]{246,223,172} -13.5 & \cellcolor[RGB]{251,241,218} -6.0 & \cellcolor[RGB]{245,219,160} -15.2 & \cellcolor[RGB]{247,226,179} -11.9 \\
Molmo-72B & -0.6 & +0.4 & -0.2 & +0.5 & +0.5 & +0.3 & +0.6 & +0.4 & \cellcolor[RGB]{246,222,168} -14.1 & -1.9 & \cellcolor[RGB]{190,219,235} +10.6 & \cellcolor[RGB]{248,231,192} -10.0 & \cellcolor[RGB]{248,230,190} -10.3 & -0.4 & \cellcolor[RGB]{232,168,26} -37.0 & \cellcolor[RGB]{232,169,28} -36.6 \\
Qwen2.5-VL-72B & \cellcolor[RGB]{252,246,232} -3.6 & \cellcolor[RGB]{253,250,242} -2.1 & \cellcolor[RGB]{253,247,236} -3.0 & \cellcolor[RGB]{206,227,240} +7.9 & \cellcolor[RGB]{216,233,243} +6.4 & -1.7 & -1.2 & -1.4 & \cellcolor[RGB]{247,226,179} -11.9 & \cellcolor[RGB]{242,208,131} -19.8 & \cellcolor[RGB]{249,232,195} -9.4 & \cellcolor[RGB]{242,247,251} +2.0 & +1.6 & \cellcolor[RGB]{242,205,124} -21.0 & \cellcolor[RGB]{245,216,154} -16.1 & \cellcolor[RGB]{245,217,156} -15.8 \\
GPT-4.1-Mini & +0.9 & \cellcolor[RGB]{240,246,250} +2.5 & \cellcolor[RGB]{236,244,249} +3.1 & \cellcolor[RGB]{179,213,232} +11.9 & \cellcolor[RGB]{184,215,233} +11.4 & +1.8 & \cellcolor[RGB]{240,246,250} +2.4 & \cellcolor[RGB]{244,248,251} +1.9 & \cellcolor[RGB]{249,232,195} -9.5 & \cellcolor[RGB]{246,223,172} -13.5 & \cellcolor[RGB]{244,216,152} -16.6 & \cellcolor[RGB]{252,244,227} -4.4 & \cellcolor[RGB]{252,244,226} -4.5 & \cellcolor[RGB]{247,224,174} -13.2 & \cellcolor[RGB]{245,219,162} -14.8 & \cellcolor[RGB]{247,224,174} -13.1 \\
\hline
MedGemma-4B & -1.3 & -0.2 & -0.9 & \cellcolor[RGB]{244,248,251} -1.9 & \cellcolor[RGB]{242,247,251} -2.0 & -0.4 & -0.4 & -0.5 & \cellcolor[RGB]{240,197,102} +24.6 & \cellcolor[RGB]{241,201,113} +22.6 & \cellcolor[RGB]{243,211,140} +18.5 & \cellcolor[RGB]{237,189,80} +28.1 & \cellcolor[RGB]{238,191,86} +27.1 & \cellcolor[RGB]{242,205,124} +21.2 & \cellcolor[RGB]{239,195,97} +25.2 & \cellcolor[RGB]{240,198,105} +24.1 \\
Qwen2-VL-7B & +0.1 & -0.2 & -0.1 & 0.0 & +0.2 & -0.3 & +1.2 & +1.0 & \cellcolor[RGB]{254,251,246} +1.3 & +0.2 & +0.5 & \cellcolor[RGB]{249,232,194} +9.7 & \cellcolor[RGB]{253,250,242} +2.0 & \cellcolor[RGB]{210,230,241} -7.3 & \cellcolor[RGB]{253,250,242} +2.1 & \cellcolor[RGB]{253,250,243} +1.8 \\
Qwen2.5-VL-7B & +1.0 & +0.6 & +0.6 & -0.6 & -0.7 & +0.5 & -0.3 & -0.2 & \cellcolor[RGB]{246,220,163} +14.7 & \cellcolor[RGB]{248,228,184} +11.5 & \cellcolor[RGB]{248,228,184} +11.5 & \cellcolor[RGB]{240,200,110} +23.2 & \cellcolor[RGB]{241,202,116} +22.5 & \cellcolor[RGB]{252,244,226} +4.7 & \cellcolor[RGB]{248,228,184} +11.5 & \cellcolor[RGB]{251,241,220} +5.7 \\
Molmo-7B & \cellcolor[RGB]{253,249,240} +2.3 & \cellcolor[RGB]{232,242,248} -3.8 & \cellcolor[RGB]{236,244,249} -3.0 & +1.0 & +1.0 & \cellcolor[RGB]{230,241,247} -4.1 & 0.0 & -0.2 & \cellcolor[RGB]{241,201,113} +22.7 & \cellcolor[RGB]{234,243,248} -3.3 & +1.0 & \cellcolor[RGB]{239,196,100} +25.0 & \cellcolor[RGB]{239,195,96} +25.5 & \cellcolor[RGB]{184,215,233} -11.4 & \cellcolor[RGB]{251,243,224} +5.0 & \cellcolor[RGB]{230,241,247} -3.9 \\
LLaVA-v1.6-7B & \cellcolor[RGB]{248,251,252} -1.1 & -0.2 & \cellcolor[RGB]{244,248,251} -1.7 & \cellcolor[RGB]{224,237,245} -5.1 & \cellcolor[RGB]{228,240,246} -4.3 & -0.4 & \cellcolor[RGB]{246,250,252} -1.6 & \cellcolor[RGB]{244,248,251} -1.8 & \cellcolor[RGB]{250,238,210} +7.4 & \cellcolor[RGB]{248,228,184} +11.4 & \cellcolor[RGB]{250,236,206} +7.8 & \cellcolor[RGB]{247,226,179} +12.2 & \cellcolor[RGB]{248,229,186} +11.2 & \cellcolor[RGB]{249,235,202} +8.6 & \cellcolor[RGB]{247,224,174} +13.2 & \cellcolor[RGB]{248,229,188} +10.9 \\
LLaVA-OV-8B & \cellcolor[RGB]{254,252,248} +1.1 & \cellcolor[RGB]{251,241,220} +5.5 & \cellcolor[RGB]{253,248,238} +2.8 & \cellcolor[RGB]{244,248,251} -1.8 & -0.5 & \cellcolor[RGB]{253,250,242} +2.0 & \cellcolor[RGB]{254,251,246} +1.6 & +1.2 & \cellcolor[RGB]{248,229,188} +10.7 & \cellcolor[RGB]{241,204,121} +21.4 & \cellcolor[RGB]{245,217,156} +15.9 & \cellcolor[RGB]{247,226,179} +12.2 & \cellcolor[RGB]{248,230,190} +10.4 & \cellcolor[RGB]{253,250,242} +2.1 & \cellcolor[RGB]{250,237,208} +7.6 & \cellcolor[RGB]{253,248,238} +2.7 \\
MedGemma-27B & +1.3 & \cellcolor[RGB]{242,247,251} -2.2 & \cellcolor[RGB]{238,245,249} -2.7 & \cellcolor[RGB]{228,240,246} -4.3 & \cellcolor[RGB]{232,242,248} -3.7 & \cellcolor[RGB]{242,247,251} -2.1 & \cellcolor[RGB]{236,244,249} -2.9 & \cellcolor[RGB]{234,243,248} -3.4 & \cellcolor[RGB]{239,194,94} +25.9 & \cellcolor[RGB]{242,247,251} -2.1 & -0.1 & \cellcolor[RGB]{247,226,179} +12.1 & \cellcolor[RGB]{248,229,186} +11.1 & \cellcolor[RGB]{214,232,242} -6.7 & +0.4 & \cellcolor[RGB]{234,243,248} -3.3 \\
Qwen2.5-VL-32B & -0.2 & \cellcolor[RGB]{240,246,250} -2.5 & \cellcolor[RGB]{242,247,251} -2.1 & \cellcolor[RGB]{240,246,250} -2.4 & \cellcolor[RGB]{244,248,251} -1.7 & \cellcolor[RGB]{246,250,252} -1.5 & \cellcolor[RGB]{248,251,252} -1.2 & \cellcolor[RGB]{240,246,250} -2.3 & \cellcolor[RGB]{246,223,172} +13.3 & \cellcolor[RGB]{251,241,218} +5.8 & \cellcolor[RGB]{251,241,220} +5.7 & \cellcolor[RGB]{244,213,144} +18.0 & \cellcolor[RGB]{244,214,147} +17.3 & \cellcolor[RGB]{252,246,232} +3.7 & \cellcolor[RGB]{249,233,198} +9.3 & \cellcolor[RGB]{253,250,242} +2.0 \\
LLaVA-v1.6-34B & -0.6 & \cellcolor[RGB]{253,250,243} +1.9 & -0.6 & \cellcolor[RGB]{236,244,249} -3.0 & \cellcolor[RGB]{242,247,251} -2.2 & +0.8 & +0.5 & +0.5 & \cellcolor[RGB]{247,226,178} +12.4 & \cellcolor[RGB]{240,199,108} +23.6 & \cellcolor[RGB]{244,213,144} +18.0 & \cellcolor[RGB]{244,215,150} +16.8 & \cellcolor[RGB]{245,217,156} +16.1 & \cellcolor[RGB]{250,235,204} +8.2 & \cellcolor[RGB]{246,223,170} +13.7 & \cellcolor[RGB]{249,232,195} +9.4 \\
Molmo-72B & \cellcolor[RGB]{244,248,251} -1.8 & 0.0 & 0.0 & -1.5 & -1.4 & 0.0 & -0.7 & 0.0 & \cellcolor[RGB]{246,223,172} +13.3 & \cellcolor[RGB]{204,226,239} -8.2 & \cellcolor[RGB]{176,211,231} -12.7 & \cellcolor[RGB]{243,211,140} +18.6 & \cellcolor[RGB]{244,215,150} +17.0 & \cellcolor[RGB]{170,208,229} -13.7 & \cellcolor[RGB]{253,247,236} +3.1 & \cellcolor[RGB]{226,238,246} -4.6 \\
Qwen2.5-VL-72B & \cellcolor[RGB]{253,249,240} +2.3 & \cellcolor[RGB]{252,247,234} +3.4 & \cellcolor[RGB]{252,247,234} +3.4 & \cellcolor[RGB]{216,233,243} -6.4 & \cellcolor[RGB]{222,236,245} -5.2 & \cellcolor[RGB]{254,251,246} +1.6 & +1.1 & +1.1 & \cellcolor[RGB]{249,233,198} +9.1 & \cellcolor[RGB]{244,216,152} +16.5 & \cellcolor[RGB]{249,232,194} +9.8 & -0.3 & -0.3 & \cellcolor[RGB]{246,221,166} +14.3 & \cellcolor[RGB]{246,220,163} +14.6 & \cellcolor[RGB]{246,222,168} +14.0 \\
GPT-4.1-Mini & \cellcolor[RGB]{253,248,238} +2.7 & \cellcolor[RGB]{253,250,242} +2.1 & +0.5 & \cellcolor[RGB]{190,219,235} -10.5 & \cellcolor[RGB]{200,224,238} -8.7 & \cellcolor[RGB]{253,250,242} +2.0 & -0.5 & -0.4 & \cellcolor[RGB]{248,230,190} +10.5 & \cellcolor[RGB]{248,228,184} +11.5 & \cellcolor[RGB]{247,226,178} +12.5 & \cellcolor[RGB]{248,231,192} +10.3 & \cellcolor[RGB]{249,232,194} +9.9 & \cellcolor[RGB]{248,228,184} +11.5 & \cellcolor[RGB]{247,227,182} +11.7 & \cellcolor[RGB]{247,226,179} +12.1 \\
\hline
MedGemma-4B & \cellcolor[RGB]{244,243,249} +3.2 & \cellcolor[RGB]{252,251,253} +1.0 & \cellcolor[RGB]{250,249,252} +1.5 & \cellcolor[RGB]{241,239,247} +4.1 & \cellcolor[RGB]{240,237,247} +4.3 & \cellcolor[RGB]{251,250,252} +1.0 & \cellcolor[RGB]{251,250,252} +1.3 & \cellcolor[RGB]{251,250,252} +1.2 & \cellcolor[RGB]{248,246,251} +2.1 & \cellcolor[RGB]{253,252,250} -0.9 & -1.0 & \cellcolor[RGB]{252,251,253} +0.9 & \cellcolor[RGB]{250,249,252} +1.3 & \cellcolor[RGB]{253,251,249} -1.1 & -0.8 & \cellcolor[RGB]{253,251,249} -1.2 \\
Qwen2-VL-7B & \cellcolor[RGB]{241,239,247} +4.2 & \cellcolor[RGB]{246,244,250} +2.9 & \cellcolor[RGB]{240,237,247} +4.4 & \cellcolor[RGB]{237,234,245} +5.3 & \cellcolor[RGB]{246,244,250} +2.8 & \cellcolor[RGB]{248,246,251} +2.0 & \cellcolor[RGB]{235,231,244} +6.0 & \cellcolor[RGB]{238,235,246} +4.9 & \cellcolor[RGB]{236,232,244} +5.5 & \cellcolor[RGB]{239,236,246} +4.5 & \cellcolor[RGB]{234,230,243} +6.3 & \cellcolor[RGB]{241,239,247} +3.9 & \cellcolor[RGB]{239,236,246} +4.6 & \cellcolor[RGB]{242,240,248} +3.8 & \cellcolor[RGB]{216,208,233} +11.8 & \cellcolor[RGB]{219,212,235} +10.8 \\
Qwen2.5-VL-7B & -0.6 & +0.4 & \cellcolor[RGB]{252,250,247} -1.4 & \cellcolor[RGB]{241,239,247} +4.0 & \cellcolor[RGB]{240,237,247} +4.3 & +0.1 & \cellcolor[RGB]{246,244,250} +2.9 & \cellcolor[RGB]{247,245,250} +2.5 & \cellcolor[RGB]{252,249,245} -1.9 & \cellcolor[RGB]{249,248,251} +1.6 & \cellcolor[RGB]{252,250,247} -1.4 & \cellcolor[RGB]{252,250,247} -1.5 & \cellcolor[RGB]{252,250,247} -1.6 & \cellcolor[RGB]{250,249,252} +1.4 & \cellcolor[RGB]{239,236,246} +4.6 & \cellcolor[RGB]{241,239,247} +4.0 \\
Molmo-7B & \cellcolor[RGB]{217,181,132} -24.1 & \cellcolor[RGB]{223,192,150} -20.6 & \cellcolor[RGB]{209,165,105} -29.2 & -0.5 & -0.5 & \cellcolor[RGB]{235,215,189} -12.8 & +0.3 & +0.6 & \cellcolor[RGB]{211,168,110} -28.3 & \cellcolor[RGB]{216,178,127} -25.0 & \cellcolor[RGB]{202,151,82} -33.8 & \cellcolor[RGB]{249,244,237} -3.5 & \cellcolor[RGB]{249,243,236} -3.6 & \cellcolor[RGB]{228,203,168} -17.0 & \cellcolor[RGB]{251,248,244} -2.1 & \cellcolor[RGB]{251,247,242} -2.5 \\
LLaVA-v1.6-7B & \cellcolor[RGB]{240,237,247} +4.5 & -0.0 & \cellcolor[RGB]{240,237,247} +4.3 & \cellcolor[RGB]{217,210,234} +11.5 & \cellcolor[RGB]{214,206,232} +12.3 & +0.0 & \cellcolor[RGB]{240,237,247} +4.5 & \cellcolor[RGB]{243,241,248} +3.3 & -0.8 & \cellcolor[RGB]{250,246,241} -2.7 & -0.4 & \cellcolor[RGB]{253,251,249} -1.1 & \cellcolor[RGB]{252,250,247} -1.4 & \cellcolor[RGB]{250,246,241} -2.7 & \cellcolor[RGB]{236,232,244} +5.6 & \cellcolor[RGB]{243,241,248} +3.4 \\
LLaVA-OV-8B & \cellcolor[RGB]{250,249,252} +1.6 & \cellcolor[RGB]{253,251,249} -1.2 & +0.2 & \cellcolor[RGB]{219,212,235} +10.7 & \cellcolor[RGB]{217,210,234} +11.3 & \cellcolor[RGB]{249,244,237} -3.3 & \cellcolor[RGB]{251,250,252} +1.2 & -1.1 & \cellcolor[RGB]{248,246,251} +2.2 & -0.2 & \cellcolor[RGB]{252,251,253} +1.0 & \cellcolor[RGB]{250,249,252} +1.5 & \cellcolor[RGB]{250,249,252} +1.4 & +0.1 & \cellcolor[RGB]{238,235,246} +4.9 & \cellcolor[RGB]{246,244,250} +2.8 \\
MedGemma-27B & +0.5 & \cellcolor[RGB]{243,241,248} +3.2 & \cellcolor[RGB]{231,226,241} +7.1 & \cellcolor[RGB]{216,208,233} +11.9 & \cellcolor[RGB]{214,206,232} +12.5 & \cellcolor[RGB]{243,241,248} +3.5 & \cellcolor[RGB]{233,229,243} +6.6 & \cellcolor[RGB]{230,225,241} +7.6 & -0.0 & \cellcolor[RGB]{240,237,247} +4.3 & \cellcolor[RGB]{224,219,238} +9.3 & \cellcolor[RGB]{234,230,243} +6.3 & \cellcolor[RGB]{233,229,243} +6.5 & \cellcolor[RGB]{239,236,246} +4.5 & \cellcolor[RGB]{235,231,244} +6.0 & \cellcolor[RGB]{230,225,241} +7.6 \\
Qwen2.5-VL-32B & \cellcolor[RGB]{243,241,248} +3.5 & \cellcolor[RGB]{218,211,234} +11.2 & \cellcolor[RGB]{212,203,231} +13.2 & \cellcolor[RGB]{230,225,241} +7.6 & \cellcolor[RGB]{229,224,240} +7.8 & \cellcolor[RGB]{216,208,233} +11.7 & \cellcolor[RGB]{242,240,248} +3.8 & \cellcolor[RGB]{234,230,243} +6.4 & \cellcolor[RGB]{248,246,251} +1.9 & \cellcolor[RGB]{224,219,238} +9.0 & \cellcolor[RGB]{220,213,236} +10.6 & \cellcolor[RGB]{251,250,252} +1.3 & \cellcolor[RGB]{251,250,252} +1.1 & \cellcolor[RGB]{222,216,237} +9.8 & \cellcolor[RGB]{240,237,247} +4.4 & \cellcolor[RGB]{235,231,244} +6.0 \\
LLaVA-v1.6-34B & \cellcolor[RGB]{235,231,244} +6.0 & -0.2 & \cellcolor[RGB]{247,245,250} +2.5 & \cellcolor[RGB]{223,217,237} +9.5 & \cellcolor[RGB]{219,212,235} +10.7 & -0.5 & \cellcolor[RGB]{250,249,252} +1.5 & -0.1 & -0.8 & \cellcolor[RGB]{252,250,247} -1.4 & \cellcolor[RGB]{253,251,249} -1.0 & \cellcolor[RGB]{252,250,247} -1.4 & \cellcolor[RGB]{252,250,247} -1.5 & \cellcolor[RGB]{248,241,232} -4.3 & -0.4 & \cellcolor[RGB]{250,245,239} -3.0 \\
Molmo-72B & \cellcolor[RGB]{238,235,246} +4.9 & +0.4 & -0.3 & \cellcolor[RGB]{239,236,246} +4.8 & \cellcolor[RGB]{237,234,245} +5.2 & +0.3 & \cellcolor[RGB]{246,244,250} +2.7 & \cellcolor[RGB]{249,248,251} +1.8 & +0.6 & -0.4 & \cellcolor[RGB]{249,243,236} -3.6 & +0.3 & +0.3 & \cellcolor[RGB]{254,253,252} -0.5 & \cellcolor[RGB]{227,222,240} +8.4 & \cellcolor[RGB]{236,232,244} +5.7 \\
Qwen2.5-VL-72B & \cellcolor[RGB]{238,235,246} +4.9 & \cellcolor[RGB]{250,249,252} +1.4 & \cellcolor[RGB]{250,249,252} +1.3 & \cellcolor[RGB]{195,183,222} +18.3 & \cellcolor[RGB]{193,181,221} +18.8 & \cellcolor[RGB]{248,246,251} +2.2 & \cellcolor[RGB]{243,241,248} +3.5 & \cellcolor[RGB]{244,243,249} +3.0 & \cellcolor[RGB]{247,245,250} +2.4 & \cellcolor[RGB]{252,249,245} -1.7 & \cellcolor[RGB]{246,244,250} +2.7 & \cellcolor[RGB]{218,211,234} +11.3 & \cellcolor[RGB]{220,213,236} +10.3 & \cellcolor[RGB]{249,244,237} -3.5 & -0.5 & \cellcolor[RGB]{252,249,245} -1.6 \\
GPT-4.1-Mini & \cellcolor[RGB]{244,243,249} +3.1 & \cellcolor[RGB]{247,245,250} +2.5 & \cellcolor[RGB]{238,235,246} +4.8 & \cellcolor[RGB]{182,168,215} +22.1 & \cellcolor[RGB]{181,167,214} +22.3 & \cellcolor[RGB]{246,244,250} +2.9 & \cellcolor[RGB]{226,221,239} +8.5 & \cellcolor[RGB]{230,225,241} +7.5 & +0.6 & \cellcolor[RGB]{251,250,252} +1.2 & \cellcolor[RGB]{243,241,248} +3.5 & +0.4 & +0.7 & \cellcolor[RGB]{250,249,252} +1.4 & \cellcolor[RGB]{231,226,241} +7.3 & \cellcolor[RGB]{235,231,244} +6.1 \\
\end{tabular}
\caption{Changes in AUROC (↑; top block), Brier (↓; middle block), and average estimated uncertainty (bottom block) for \texttt{+nota} and \texttt{-correct+nota} experiments per model and per UE method, measured relative to the \texttt{original} benchmark (Table~\ref{tab:main_table}). Blue/orange shading indicates significantly higher/lower performance; purple/brown indicates significantly higher/lower average uncertainty.}
\label{app_tab:deltas_all_ue}
\end{table*}

%% file: nota_averages_multiple_models_appendix_small.tex
\begin{table}
    
\centering
\scriptsize
\setlength{\tabcolsep}{3pt}
\begin{tabular}{r | cccc | cccc}

\textbf{UE} & \multicolumn{4}{c}{\texttt{+nota}} & \multicolumn{4}{c}{\texttt{-correct+nota}} \\

\textbf{Method} & \CtoC & \CtoW  &
\WtoC  &
\WtoW  & \CtoC & \CtoW  & 
\WtoC  &
\WtoW  \\
\midrule
\multicolumn{9}{l}{MedGemma-4B}\\
\midrule
LabelNLL & \cellcolor[RGB]{103,171,209} -0.35 & \cellcolor[RGB]{238,191,87} +2.16 & \cellcolor[RGB]{240,200,109} +1.63 & \cellcolor[RGB]{164,205,227} -0.13 & \cellcolor[RGB]{16,122,182} -0.88 & +0.10 & -0.00 & +0.00 \\
ANLL & \cellcolor[RGB]{171,208,229} -0.11 & \cellcolor[RGB]{245,220,162} +0.66 & \cellcolor[RGB]{246,223,170} +0.55 & \cellcolor[RGB]{202,226,239} -0.04 & \cellcolor[RGB]{73,154,200} -0.51 & \cellcolor[RGB]{252,244,226} +0.06 & -0.16 & +0.00 \\
MaxNLL & \cellcolor[RGB]{177,212,231} -0.09 & \cellcolor[RGB]{246,222,168} +0.57 & \cellcolor[RGB]{247,225,175} +0.48 & -0.04 & \cellcolor[RGB]{83,159,203} -0.45 & +0.05 & -0.23 & +0.00 \\
SC & \cellcolor[RGB]{104,171,209} -0.35 & \cellcolor[RGB]{238,192,88} +2.13 & \cellcolor[RGB]{239,197,101} +1.82 & \cellcolor[RGB]{159,202,226} -0.14 & \cellcolor[RGB]{6,117,180} -0.95 & +0.11 & -0.46 & +0.00 \\
SE & \cellcolor[RGB]{98,168,207} -0.38 & \cellcolor[RGB]{237,189,81} +2.32 & \cellcolor[RGB]{239,196,100} +1.83 & \cellcolor[RGB]{159,202,226} -0.14 & \cellcolor[RGB]{8,118,180} -0.94 & +0.11 & -0.36 & +0.00 \\
PRO & \cellcolor[RGB]{173,209,230} -0.10 & \cellcolor[RGB]{246,221,164} +0.63 & \cellcolor[RGB]{246,223,172} +0.53 & \cellcolor[RGB]{203,226,239} -0.04 & \cellcolor[RGB]{76,156,201} -0.49 & \cellcolor[RGB]{252,244,227} +0.06 & -0.15 & +0.00 \\
EE & \cellcolor[RGB]{174,210,230} -0.10 & \cellcolor[RGB]{246,221,165} +0.62 & \cellcolor[RGB]{247,226,178} +0.45 & -0.03 & \cellcolor[RGB]{57,145,195} -0.60 & +0.07 & -0.28 & +0.00 \\
RDS & \cellcolor[RGB]{173,209,230} -0.10 & \cellcolor[RGB]{246,220,164} +0.63 & \cellcolor[RGB]{247,226,178} +0.46 & -0.04 & \cellcolor[RGB]{57,145,195} -0.60 & +0.07 & -0.34 & +0.00 \\
\hline
Num. & 859 & 270 & 230 & 1259 & 140 & 881 & 32 & 1309 \\
\midrule
\multicolumn{9}{l}{Qwen2-VL-7B}\\
\midrule
LabelNLL & \cellcolor[RGB]{164,205,227} -0.13 & \cellcolor[RGB]{246,222,168} +0.58 & \cellcolor[RGB]{249,235,202} +0.22 & \cellcolor[RGB]{207,228,240} -0.03 & \cellcolor[RGB]{85,161,203} -0.44 & \cellcolor[RGB]{251,240,216} +0.11 & \cellcolor[RGB]{249,233,196} +0.26 & -0.00 \\
ANLL & \cellcolor[RGB]{167,206,228} -0.12 & \cellcolor[RGB]{246,223,170} +0.54 & \cellcolor[RGB]{249,235,202} +0.21 & \cellcolor[RGB]{208,229,240} -0.03 & \cellcolor[RGB]{85,161,203} -0.44 & \cellcolor[RGB]{251,240,216} +0.11 & \cellcolor[RGB]{249,232,194} +0.28 & -0.00 \\
MaxNLL & \cellcolor[RGB]{166,205,228} -0.12 & \cellcolor[RGB]{246,222,169} +0.56 & \cellcolor[RGB]{249,235,202} +0.21 & \cellcolor[RGB]{207,229,240} -0.03 & \cellcolor[RGB]{85,161,203} -0.44 & \cellcolor[RGB]{251,240,216} +0.11 & \cellcolor[RGB]{249,232,195} +0.28 & -0.00 \\
SC & \cellcolor[RGB]{130,186,217} -0.24 & \cellcolor[RGB]{243,209,135} +1.10 & \cellcolor[RGB]{247,227,181} +0.41 & \cellcolor[RGB]{189,218,235} -0.07 & \cellcolor[RGB]{48,140,192} -0.66 & \cellcolor[RGB]{250,237,207} +0.17 & \cellcolor[RGB]{248,228,184} +0.39 & -0.01 \\
SE & \cellcolor[RGB]{136,189,219} -0.22 & \cellcolor[RGB]{243,212,140} +1.00 & \cellcolor[RGB]{248,229,187} +0.35 & \cellcolor[RGB]{194,221,236} -0.06 & \cellcolor[RGB]{49,141,192} -0.65 & \cellcolor[RGB]{250,237,208} +0.17 & \cellcolor[RGB]{247,227,181} +0.42 & -0.01 \\
PRO & \cellcolor[RGB]{179,213,232} -0.09 & \cellcolor[RGB]{247,227,182} +0.41 & \cellcolor[RGB]{250,238,211} +0.15 & \cellcolor[RGB]{216,233,243} -0.02 & \cellcolor[RGB]{102,170,209} -0.36 & \cellcolor[RGB]{251,241,220} +0.09 & \cellcolor[RGB]{250,238,211} +0.15 & -0.00 \\
EE & \cellcolor[RGB]{150,197,223} -0.17 & \cellcolor[RGB]{245,217,154} +0.78 & \cellcolor[RGB]{248,228,185} +0.37 & \cellcolor[RGB]{193,220,236} -0.06 & \cellcolor[RGB]{54,144,194} -0.62 & \cellcolor[RGB]{250,237,209} +0.16 & \cellcolor[RGB]{246,220,163} +0.64 & -0.01 \\
RDS & \cellcolor[RGB]{146,194,222} -0.18 & \cellcolor[RGB]{244,215,150} +0.84 & \cellcolor[RGB]{248,230,189} +0.33 & \cellcolor[RGB]{196,222,237} -0.05 & \cellcolor[RGB]{53,143,194} -0.62 & \cellcolor[RGB]{250,237,209} +0.16 & \cellcolor[RGB]{247,226,179} +0.44 & -0.01 \\
\hline
Num. & 863 & 336 & 353 & 1091 & 274 & 890 & 42 & 1189 \\
\midrule
\multicolumn{9}{l}{Qwen2.5-VL-7B}\\
\midrule
LabelNLL & \cellcolor[RGB]{158,201,225} -0.14 & \cellcolor[RGB]{244,214,148} +0.87 & \cellcolor[RGB]{248,228,183} +0.39 & \cellcolor[RGB]{199,224,238} -0.05 & \cellcolor[RGB]{67,151,198} -0.54 & \cellcolor[RGB]{251,243,223} +0.07 & \cellcolor[RGB]{248,229,186} +0.36 & -0.00 \\
ANLL & -0.03 & \cellcolor[RGB]{250,236,205} +0.19 & +0.02 & -0.00 & \cellcolor[RGB]{157,200,225} -0.15 & +0.02 & -0.12 & +0.00 \\
MaxNLL & -0.03 & \cellcolor[RGB]{250,236,204} +0.20 & +0.05 & -0.01 & \cellcolor[RGB]{154,199,224} -0.16 & +0.02 & -0.25 & +0.00 \\
SC & \cellcolor[RGB]{121,180,214} -0.28 & \cellcolor[RGB]{240,199,107} +1.66 & \cellcolor[RGB]{245,217,155} +0.77 & \cellcolor[RGB]{177,211,231} -0.09 & \cellcolor[RGB]{22,126,184} -0.83 & \cellcolor[RGB]{251,240,216} +0.11 & +0.49 & -0.00 \\
SE & \cellcolor[RGB]{126,183,216} -0.25 & \cellcolor[RGB]{241,201,113} +1.53 & \cellcolor[RGB]{245,220,162} +0.66 & \cellcolor[RGB]{182,215,233} -0.08 & \cellcolor[RGB]{25,127,185} -0.81 & \cellcolor[RGB]{251,240,216} +0.11 & +0.47 & -0.00 \\
PRO & \cellcolor[RGB]{196,222,237} -0.05 & \cellcolor[RGB]{246,222,167} +0.59 & \cellcolor[RGB]{247,227,182} +0.40 & -0.03 & \cellcolor[RGB]{140,191,220} -0.20 & +0.04 & \cellcolor[RGB]{246,221,166} +0.60 & -0.01 \\
EE & \cellcolor[RGB]{169,207,229} -0.11 & \cellcolor[RGB]{245,219,161} +0.68 & \cellcolor[RGB]{248,228,184} +0.38 & \cellcolor[RGB]{200,224,238} -0.05 & \cellcolor[RGB]{91,164,205} -0.41 & \cellcolor[RGB]{252,244,227} +0.06 & +0.15 & -0.00 \\
RDS & \cellcolor[RGB]{174,210,230} -0.10 & \cellcolor[RGB]{246,221,166} +0.60 & \cellcolor[RGB]{248,229,187} +0.35 & \cellcolor[RGB]{202,225,239} -0.04 & \cellcolor[RGB]{96,167,207} -0.39 & \cellcolor[RGB]{252,245,228} +0.05 & +0.17 & -0.00 \\
\hline
Num. & 849 & 301 & 312 & 1115 & 162 & 913 & 15 & 1203 \\
\midrule
\multicolumn{9}{l}{Molmo-7B}\\
\midrule
LabelNLL & \cellcolor[RGB]{195,221,236} -0.05 & \cellcolor[RGB]{248,230,190} +0.32 & \cellcolor[RGB]{248,229,187} +0.35 & \cellcolor[RGB]{211,230,241} -0.03 & \cellcolor[RGB]{100,169,208} -0.37 & +0.01 & +0.52 & -0.00 \\
ANLL & -0.01 & \cellcolor[RGB]{252,243,225} +0.07 & \cellcolor[RGB]{251,242,222} +0.08 & -0.01 & -0.05 & +0.00 & +0.08 & -0.00 \\
MaxNLL & -0.01 & +0.04 & +0.05 & -0.00 & +0.08 & -0.00 & -0.03 & +0.00 \\
SC & \cellcolor[RGB]{125,183,215} -0.26 & \cellcolor[RGB]{241,202,115} +1.50 & \cellcolor[RGB]{244,212,143} +0.96 & \cellcolor[RGB]{182,215,233} -0.08 & \cellcolor[RGB]{39,135,189} -0.72 & +0.02 & \cellcolor[RGB]{243,208,132} +1.15 & -0.00 \\
SE & \cellcolor[RGB]{132,187,217} -0.23 & \cellcolor[RGB]{242,205,122} +1.34 & \cellcolor[RGB]{244,215,149} +0.86 & \cellcolor[RGB]{186,217,234} -0.07 & \cellcolor[RGB]{45,139,191} -0.67 & +0.02 & \cellcolor[RGB]{242,207,128} +1.23 & -0.00 \\
PRO & -0.01 & \cellcolor[RGB]{252,244,228} +0.05 & \cellcolor[RGB]{252,246,232} +0.04 & -0.00 & -0.07 & +0.00 & +0.09 & -0.00 \\
EE & \cellcolor[RGB]{201,225,238} -0.04 & \cellcolor[RGB]{249,233,197} +0.26 & \cellcolor[RGB]{250,236,205} +0.19 & -0.02 & -0.06 & +0.00 & +0.02 & -0.00 \\
RDS & \cellcolor[RGB]{210,230,241} -0.03 & \cellcolor[RGB]{250,236,206} +0.18 & \cellcolor[RGB]{250,239,213} +0.13 & -0.01 & -0.03 & +0.00 & -0.01 & +0.00 \\
\hline
Num. & 806 & 281 & 294 & 1319 & 37 & 887 & 5 & 1380 \\
\midrule
\multicolumn{9}{l}{LLaVA-v1.6-7B}\\
\midrule
LabelNLL & \cellcolor[RGB]{148,196,222} -0.17 & \cellcolor[RGB]{244,214,148} +0.87 & \cellcolor[RGB]{247,227,182} +0.41 & \cellcolor[RGB]{204,227,239} -0.04 & \cellcolor[RGB]{61,147,196} -0.58 & \cellcolor[RGB]{251,241,218} +0.10 & +0.15 & -0.00 \\
ANLL & \cellcolor[RGB]{168,207,228} -0.11 & \cellcolor[RGB]{246,222,168} +0.58 & \cellcolor[RGB]{249,234,199} +0.24 & \cellcolor[RGB]{216,233,243} -0.02 & \cellcolor[RGB]{100,169,208} -0.37 & \cellcolor[RGB]{252,243,225} +0.07 & \cellcolor[RGB]{250,237,208} +0.16 & -0.00 \\
MaxNLL & \cellcolor[RGB]{160,202,226} -0.14 & \cellcolor[RGB]{245,219,160} +0.69 & \cellcolor[RGB]{248,229,186} +0.36 & \cellcolor[RGB]{207,228,240} -0.03 & \cellcolor[RGB]{97,167,207} -0.38 & \cellcolor[RGB]{252,243,225} +0.07 & +0.12 & -0.00 \\
SC & \cellcolor[RGB]{125,183,215} -0.26 & \cellcolor[RGB]{242,206,125} +1.30 & \cellcolor[RGB]{245,218,157} +0.73 & \cellcolor[RGB]{187,217,234} -0.07 & \cellcolor[RGB]{49,141,193} -0.65 & \cellcolor[RGB]{251,240,216} +0.12 & \cellcolor[RGB]{248,231,191} +0.31 & -0.01 \\
SE & \cellcolor[RGB]{130,186,217} -0.24 & \cellcolor[RGB]{242,207,130} +1.20 & \cellcolor[RGB]{246,221,165} +0.61 & \cellcolor[RGB]{193,220,236} -0.06 & \cellcolor[RGB]{51,142,193} -0.64 & \cellcolor[RGB]{251,240,216} +0.11 & +0.18 & -0.00 \\
PRO & \cellcolor[RGB]{174,210,230} -0.10 & \cellcolor[RGB]{247,224,174} +0.50 & \cellcolor[RGB]{249,234,201} +0.22 & \cellcolor[RGB]{217,234,243} -0.02 & \cellcolor[RGB]{107,173,210} -0.33 & \cellcolor[RGB]{252,244,227} +0.06 & \cellcolor[RGB]{250,237,209} +0.16 & -0.00 \\
EE & \cellcolor[RGB]{148,196,222} -0.17 & \cellcolor[RGB]{244,214,147} +0.88 & \cellcolor[RGB]{247,225,177} +0.47 & \cellcolor[RGB]{201,225,238} -0.04 & \cellcolor[RGB]{91,164,205} -0.41 & \cellcolor[RGB]{251,243,224} +0.07 & \cellcolor[RGB]{249,233,197} +0.25 & -0.00 \\
RDS & \cellcolor[RGB]{151,197,223} -0.17 & \cellcolor[RGB]{244,215,150} +0.84 & \cellcolor[RGB]{247,225,177} +0.46 & \cellcolor[RGB]{201,225,238} -0.04 & \cellcolor[RGB]{94,166,206} -0.40 & \cellcolor[RGB]{252,243,224} +0.07 & \cellcolor[RGB]{249,235,203} +0.21 & -0.00 \\
\hline
Num. & 686 & 264 & 240 & 1083 & 163 & 736 & 49 & 1119 \\
\midrule
\multicolumn{9}{l}{LLaVA-OV-8B}\\
\midrule
LabelNLL & \cellcolor[RGB]{136,189,219} -0.22 & \cellcolor[RGB]{245,217,154} +0.77 & \cellcolor[RGB]{249,232,195} +0.28 & \cellcolor[RGB]{200,224,238} -0.05 & \cellcolor[RGB]{68,151,198} -0.53 & \cellcolor[RGB]{251,239,214} +0.13 & \cellcolor[RGB]{247,226,178} +0.46 & -0.01 \\
ANLL & -0.06 & \cellcolor[RGB]{249,235,202} +0.21 & +0.03 & -0.00 & -0.09 & +0.02 & +0.27 & -0.00 \\
MaxNLL & \cellcolor[RGB]{188,218,234} -0.07 & \cellcolor[RGB]{249,233,198} +0.24 & \cellcolor[RGB]{251,241,218} +0.10 & -0.02 & \cellcolor[RGB]{158,201,225} -0.14 & +0.03 & +0.19 & -0.00 \\
SC & \cellcolor[RGB]{107,173,210} -0.33 & \cellcolor[RGB]{242,208,130} +1.19 & \cellcolor[RGB]{247,225,176} +0.47 & \cellcolor[RGB]{183,215,233} -0.08 & \cellcolor[RGB]{43,138,191} -0.69 & \cellcolor[RGB]{250,237,209} +0.16 & \cellcolor[RGB]{247,224,173} +0.51 & -0.01 \\
SE & \cellcolor[RGB]{113,176,212} -0.31 & \cellcolor[RGB]{243,210,135} +1.10 & \cellcolor[RGB]{247,228,183} +0.39 & \cellcolor[RGB]{189,218,235} -0.07 & \cellcolor[RGB]{44,138,191} -0.68 & \cellcolor[RGB]{250,237,209} +0.16 & \cellcolor[RGB]{246,223,170} +0.55 & -0.01 \\
PRO & \cellcolor[RGB]{180,214,232} -0.08 & \cellcolor[RGB]{248,231,192} +0.30 & \cellcolor[RGB]{251,241,218} +0.10 & -0.02 & \cellcolor[RGB]{135,188,218} -0.22 & \cellcolor[RGB]{252,245,228} +0.05 & +0.13 & -0.00 \\
EE & \cellcolor[RGB]{155,199,224} -0.15 & \cellcolor[RGB]{246,223,170} +0.54 & \cellcolor[RGB]{249,232,194} +0.28 & \cellcolor[RGB]{199,224,238} -0.05 & \cellcolor[RGB]{104,171,209} -0.35 & \cellcolor[RGB]{251,242,222} +0.08 & \cellcolor[RGB]{246,221,166} +0.61 & -0.01 \\
RDS & \cellcolor[RGB]{159,202,226} -0.14 & \cellcolor[RGB]{247,224,174} +0.50 & \cellcolor[RGB]{249,233,196} +0.26 & \cellcolor[RGB]{201,225,238} -0.04 & \cellcolor[RGB]{109,174,211} -0.33 & \cellcolor[RGB]{251,243,223} +0.08 & \cellcolor[RGB]{247,225,177} +0.46 & -0.01 \\
\hline
Num. & 622 & 287 & 244 & 849 & 178 & 684 & 34 & 935 \\
\end{tabular}
\caption{Relative differences (4B - 8B models) in initial average uncertainty estimates for subsets of samples where the model answer is stable vs. changes after input perturbations. Shading indicates statistical significance: blue (lower uncertainty), orange (higher uncertainty).}
\label{tab:5_2_small_models}
\end{table}

%% file: nota_averages_multiple_models_appendix_large.tex
\begin{table}
    
\centering
\scriptsize
\setlength{\tabcolsep}{3pt}
\begin{tabular}{r | cccc | cccc}

\textbf{UE} & \multicolumn{4}{c}{\texttt{+nota}} & \multicolumn{4}{c}{\texttt{-correct+nota}} \\

\textbf{Method} & \CtoC & \CtoW  &
\WtoC  &
\WtoW  & \CtoC & \CtoW  & 
\WtoC  &
\WtoW  \\
\midrule
\multicolumn{9}{l}{Qwen2.5-VL-32B}\\
\midrule
LabelNLL & \cellcolor[RGB]{139,191,220} -0.20 & \cellcolor[RGB]{243,209,134} +1.12 & \cellcolor[RGB]{248,228,183} +0.39 & \cellcolor[RGB]{196,222,237} -0.05 & \cellcolor[RGB]{72,154,199} -0.51 & \cellcolor[RGB]{250,237,207} +0.17 & \cellcolor[RGB]{246,223,170} +0.55 & -0.01 \\
ANLL & \cellcolor[RGB]{188,218,234} -0.07 & \cellcolor[RGB]{248,228,185} +0.37 & \cellcolor[RGB]{251,241,217} +0.11 & -0.01 & \cellcolor[RGB]{152,198,224} -0.16 & \cellcolor[RGB]{252,245,228} +0.05 & +0.03 & -0.00 \\
MaxNLL & \cellcolor[RGB]{192,220,236} -0.06 & \cellcolor[RGB]{248,230,189} +0.33 & \cellcolor[RGB]{251,240,217} +0.11 & -0.01 & \cellcolor[RGB]{162,203,227} -0.13 & \cellcolor[RGB]{252,245,231} +0.04 & +0.08 & -0.00 \\
SC & \cellcolor[RGB]{115,177,212} -0.30 & \cellcolor[RGB]{240,200,109} +1.64 & \cellcolor[RGB]{246,220,164} +0.63 & \cellcolor[RGB]{180,213,232} -0.09 & \cellcolor[RGB]{54,143,194} -0.62 & \cellcolor[RGB]{249,235,202} +0.21 & \cellcolor[RGB]{244,216,152} +0.82 & -0.02 \\
SE & \cellcolor[RGB]{122,181,215} -0.27 & \cellcolor[RGB]{241,202,116} +1.48 & \cellcolor[RGB]{246,222,168} +0.57 & \cellcolor[RGB]{184,215,233} -0.08 & \cellcolor[RGB]{56,145,195} -0.61 & \cellcolor[RGB]{249,235,203} +0.20 & \cellcolor[RGB]{245,217,154} +0.77 & -0.02 \\
PRO & \cellcolor[RGB]{187,217,234} -0.07 & \cellcolor[RGB]{248,228,184} +0.38 & \cellcolor[RGB]{251,240,216} +0.11 & -0.02 & \cellcolor[RGB]{149,196,223} -0.17 & \cellcolor[RGB]{252,244,227} +0.06 & +0.06 & -0.00 \\
EE & \cellcolor[RGB]{164,204,227} -0.13 & \cellcolor[RGB]{245,219,160} +0.69 & \cellcolor[RGB]{248,230,189} +0.33 & \cellcolor[RGB]{201,225,238} -0.04 & \cellcolor[RGB]{118,179,213} -0.29 & \cellcolor[RGB]{251,241,219} +0.10 & \cellcolor[RGB]{247,224,174} +0.50 & -0.01 \\
RDS & \cellcolor[RGB]{178,212,231} -0.09 & \cellcolor[RGB]{247,224,174} +0.49 & \cellcolor[RGB]{249,233,198} +0.25 & \cellcolor[RGB]{208,229,241} -0.03 & \cellcolor[RGB]{137,189,219} -0.21 & \cellcolor[RGB]{252,243,224} +0.07 & \cellcolor[RGB]{248,231,191} +0.31 & -0.01 \\
\hline
Num. & 890 & 320 & 305 & 1074 & 350 & 846 & 48 & 1136 \\
\midrule
\multicolumn{9}{l}{LLaVA-v1.6-34B}\\
\midrule
LabelNLL & \cellcolor[RGB]{152,198,224} -0.16 & \cellcolor[RGB]{242,208,132} +1.16 & \cellcolor[RGB]{247,226,179} +0.44 & \cellcolor[RGB]{203,226,239} -0.04 & \cellcolor[RGB]{28,129,186} -0.79 & \cellcolor[RGB]{250,239,213} +0.13 & \cellcolor[RGB]{248,228,184} +0.38 & -0.00 \\
ANLL & \cellcolor[RGB]{175,210,230} -0.10 & \cellcolor[RGB]{245,219,159} +0.70 & \cellcolor[RGB]{248,228,185} +0.37 & \cellcolor[RGB]{207,228,240} -0.03 & \cellcolor[RGB]{92,165,205} -0.41 & \cellcolor[RGB]{252,243,225} +0.07 & \cellcolor[RGB]{247,228,183} +0.39 & -0.00 \\
MaxNLL & \cellcolor[RGB]{169,207,229} -0.11 & \cellcolor[RGB]{244,216,152} +0.81 & \cellcolor[RGB]{248,228,184} +0.38 & \cellcolor[RGB]{206,228,240} -0.04 & \cellcolor[RGB]{89,163,204} -0.42 & \cellcolor[RGB]{252,243,224} +0.07 & \cellcolor[RGB]{248,230,190} +0.32 & -0.00 \\
SC & \cellcolor[RGB]{128,184,216} -0.25 & \cellcolor[RGB]{240,197,103} +1.77 & \cellcolor[RGB]{245,216,153} +0.80 & \cellcolor[RGB]{184,216,233} -0.08 & \cellcolor[RGB]{20,125,184} -0.85 & \cellcolor[RGB]{250,238,211} +0.14 & +0.33 & -0.00 \\
SE & \cellcolor[RGB]{134,188,218} -0.22 & \cellcolor[RGB]{240,200,110} +1.61 & \cellcolor[RGB]{245,219,161} +0.67 & \cellcolor[RGB]{190,219,235} -0.06 & \cellcolor[RGB]{20,125,184} -0.85 & \cellcolor[RGB]{250,238,211} +0.14 & \cellcolor[RGB]{247,227,182} +0.40 & -0.00 \\
PRO & \cellcolor[RGB]{180,213,232} -0.08 & \cellcolor[RGB]{246,221,166} +0.61 & \cellcolor[RGB]{248,231,191} +0.31 & \cellcolor[RGB]{211,230,241} -0.03 & \cellcolor[RGB]{97,168,207} -0.38 & \cellcolor[RGB]{252,244,226} +0.06 & \cellcolor[RGB]{248,230,190} +0.32 & -0.00 \\
EE & \cellcolor[RGB]{171,208,229} -0.11 & \cellcolor[RGB]{245,217,154} +0.77 & \cellcolor[RGB]{247,225,175} +0.48 & \cellcolor[RGB]{200,224,238} -0.05 & \cellcolor[RGB]{120,180,214} -0.28 & \cellcolor[RGB]{252,245,230} +0.05 & +0.29 & -0.00 \\
RDS & \cellcolor[RGB]{175,211,231} -0.10 & \cellcolor[RGB]{245,219,160} +0.69 & \cellcolor[RGB]{247,226,178} +0.44 & \cellcolor[RGB]{202,226,239} -0.04 & \cellcolor[RGB]{130,185,217} -0.24 & \cellcolor[RGB]{252,246,231} +0.04 & \cellcolor[RGB]{249,232,195} +0.27 & -0.00 \\
\hline
Num. & 713 & 212 & 224 & 1032 & 141 & 758 & 22 & 1075 \\
\midrule
\multicolumn{9}{l}{Molmo-72B}\\
\midrule
LabelNLL & \cellcolor[RGB]{148,195,222} -0.18 & \cellcolor[RGB]{241,203,119} +1.42 & \cellcolor[RGB]{244,214,148} +0.87 & \cellcolor[RGB]{189,219,235} -0.07 & \cellcolor[RGB]{40,136,190} -0.71 & \cellcolor[RGB]{250,239,213} +0.13 & \cellcolor[RGB]{246,222,169} +0.56 & -0.01 \\
ANLL & -0.01 & +0.05 & \cellcolor[RGB]{251,240,215} +0.12 & -0.01 & +0.05 & -0.01 & +0.05 & -0.00 \\
MaxNLL & -0.00 & +0.03 & \cellcolor[RGB]{251,242,221} +0.08 & -0.01 & +0.04 & -0.01 & +0.02 & -0.00 \\
SC & \cellcolor[RGB]{122,181,214} -0.27 & \cellcolor[RGB]{238,191,85} +2.20 & \cellcolor[RGB]{241,202,116} +1.48 & \cellcolor[RGB]{170,208,229} -0.11 & \cellcolor[RGB]{32,131,187} -0.76 & \cellcolor[RGB]{250,238,212} +0.14 & \cellcolor[RGB]{245,220,162} +0.65 & -0.01 \\
SE & \cellcolor[RGB]{129,185,216} -0.24 & \cellcolor[RGB]{239,194,95} +1.97 & \cellcolor[RGB]{242,206,126} +1.28 & \cellcolor[RGB]{176,211,231} -0.10 & \cellcolor[RGB]{33,132,188} -0.76 & \cellcolor[RGB]{250,238,212} +0.14 & \cellcolor[RGB]{245,220,163} +0.65 & -0.01 \\
PRO & -0.00 & +0.00 & \cellcolor[RGB]{214,232,242} -0.03 & +0.00 & -0.00 & +0.00 & -0.02 & +0.00 \\
EE & \cellcolor[RGB]{192,220,236} -0.06 & \cellcolor[RGB]{247,225,176} +0.48 & \cellcolor[RGB]{248,229,188} +0.35 & \cellcolor[RGB]{214,232,242} -0.03 & \cellcolor[RGB]{171,208,229} -0.11 & +0.02 & \cellcolor[RGB]{248,229,188} +0.34 & -0.00 \\
RDS & \cellcolor[RGB]{209,229,241} -0.03 & \cellcolor[RGB]{249,233,197} +0.25 & \cellcolor[RGB]{250,236,205} +0.19 & \cellcolor[RGB]{224,238,245} -0.01 & \cellcolor[RGB]{192,220,236} -0.06 & +0.01 & \cellcolor[RGB]{250,237,207} +0.18 & -0.00 \\
\hline
Num. & 824 & 212 & 230 & 1263 & 197 & 827 & 40 & 1321 \\
\midrule
\multicolumn{9}{l}{Qwen2.5-VL-72B}\\
\midrule
LabelNLL & \cellcolor[RGB]{143,193,221} -0.19 & \cellcolor[RGB]{236,183,65} +2.75 & \cellcolor[RGB]{246,220,163} +0.64 & \cellcolor[RGB]{191,219,235} -0.06 & \cellcolor[RGB]{35,133,188} -0.74 & \cellcolor[RGB]{247,224,174} +0.49 & \cellcolor[RGB]{247,226,180} +0.43 & -0.01 \\
ANLL & -0.01 & +0.16 & \cellcolor[RGB]{248,230,190} +0.32 & -0.03 & \cellcolor[RGB]{174,210,230} -0.10 & +0.07 & +0.03 & -0.00 \\
MaxNLL & -0.02 & \cellcolor[RGB]{249,234,200} +0.23 & \cellcolor[RGB]{249,234,200} +0.23 & -0.02 & \cellcolor[RGB]{169,207,229} -0.11 & \cellcolor[RGB]{251,243,223} +0.07 & +0.02 & -0.00 \\
SC & \cellcolor[RGB]{113,176,212} -0.31 & \cellcolor[RGB]{231,164,15} +4.41 & \cellcolor[RGB]{243,211,139} +1.03 & \cellcolor[RGB]{174,210,230} -0.10 & \cellcolor[RGB]{25,127,185} -0.81 & \cellcolor[RGB]{246,223,171} +0.54 & \cellcolor[RGB]{245,218,158} +0.71 & -0.02 \\
SE & \cellcolor[RGB]{122,181,214} -0.27 & \cellcolor[RGB]{232,170,29} +3.90 & \cellcolor[RGB]{244,213,144} +0.93 & \cellcolor[RGB]{178,212,231} -0.09 & \cellcolor[RGB]{26,128,185} -0.80 & \cellcolor[RGB]{246,223,171} +0.54 & \cellcolor[RGB]{246,221,164} +0.63 & -0.01 \\
PRO & -0.10 & \cellcolor[RGB]{239,197,101} +1.82 & \cellcolor[RGB]{244,213,143} +0.95 & -0.05 & \cellcolor[RGB]{96,167,207} -0.39 & \cellcolor[RGB]{248,230,190} +0.32 & \cellcolor[RGB]{245,219,159} +0.70 & -0.02 \\
EE & \cellcolor[RGB]{168,206,228} -0.12 & \cellcolor[RGB]{240,199,107} +1.67 & \cellcolor[RGB]{246,222,169} +0.56 & -0.05 & \cellcolor[RGB]{109,174,210} -0.33 & \cellcolor[RGB]{249,234,201} +0.22 & +0.28 & -0.01 \\
RDS & \cellcolor[RGB]{172,209,229} -0.11 & \cellcolor[RGB]{241,202,114} +1.52 & \cellcolor[RGB]{246,223,171} +0.53 & \cellcolor[RGB]{196,222,237} -0.05 & \cellcolor[RGB]{113,177,212} -0.31 & \cellcolor[RGB]{249,235,203} +0.20 & +0.32 & -0.01 \\
\hline
Num. & 516 & 82 & 94 & 492 & 243 & 384 & 25 & 521 \\
\midrule
\multicolumn{9}{l}{\texttt{GPT-4.1-Mini}}\\
\midrule
LabelNLL & \cellcolor[RGB]{132,187,218} -0.23 & \cellcolor[RGB]{242,206,125} +1.29 & \cellcolor[RGB]{246,222,169} +0.56 & \cellcolor[RGB]{176,211,231} -0.10 & \cellcolor[RGB]{70,152,199} -0.53 & \cellcolor[RGB]{248,230,190} +0.32 & \cellcolor[RGB]{247,225,177} +0.47 & -0.03 \\
ANLL & \cellcolor[RGB]{142,192,221} -0.19 & \cellcolor[RGB]{243,210,136} +1.08 & \cellcolor[RGB]{247,226,178} +0.45 & \cellcolor[RGB]{184,216,233} -0.08 & \cellcolor[RGB]{84,160,203} -0.45 & \cellcolor[RGB]{249,232,195} +0.27 & \cellcolor[RGB]{246,223,170} +0.55 & -0.04 \\
MaxNLL & \cellcolor[RGB]{141,192,220} -0.20 & \cellcolor[RGB]{243,209,134} +1.11 & \cellcolor[RGB]{247,225,175} +0.48 & \cellcolor[RGB]{181,214,232} -0.08 & \cellcolor[RGB]{86,161,203} -0.44 & \cellcolor[RGB]{249,232,196} +0.27 & \cellcolor[RGB]{247,224,174} +0.50 & -0.04 \\
SC & \cellcolor[RGB]{134,188,218} -0.22 & \cellcolor[RGB]{242,207,127} +1.25 & \cellcolor[RGB]{245,218,156} +0.74 & \cellcolor[RGB]{164,204,227} -0.13 & \cellcolor[RGB]{76,156,201} -0.49 & \cellcolor[RGB]{248,231,192} +0.30 & \cellcolor[RGB]{247,225,176} +0.48 & -0.03 \\
SE & \cellcolor[RGB]{135,189,219} -0.22 & \cellcolor[RGB]{242,207,129} +1.22 & \cellcolor[RGB]{245,219,160} +0.68 & \cellcolor[RGB]{168,206,228} -0.12 & \cellcolor[RGB]{73,154,200} -0.51 & \cellcolor[RGB]{248,231,191} +0.31 & \cellcolor[RGB]{247,225,177} +0.46 & -0.03 \\
PRO & \cellcolor[RGB]{143,193,221} -0.19 & \cellcolor[RGB]{243,210,136} +1.07 & \cellcolor[RGB]{247,225,177} +0.46 & \cellcolor[RGB]{183,215,233} -0.08 & \cellcolor[RGB]{85,161,203} -0.44 & \cellcolor[RGB]{249,232,196} +0.27 & \cellcolor[RGB]{246,223,171} +0.54 & -0.04 \\
EE & -0.01 & +0.05 & -0.08 & +0.01 & +0.03 & -0.02 & -0.12 & +0.01 \\
RDS & -0.01 & +0.04 & -0.06 & +0.01 & +0.03 & -0.02 & -0.14 & +0.01 \\
\hline
Num. & 740 & 237 & 277 & 848 & 390 & 606 & 139 & 916 \\
\end{tabular}
\caption{Relative differences ($>$ 32B models) in initial average uncertainty estimates for subsets of samples where the model answer is stable vs. changes after input perturbations. Shading indicates statistical significance: blue (lower uncertainty), orange (higher uncertainty).}
\label{tab:5_2_large_models}
\end{table}

%% file: nota_thresholding_multiple_models_appendix_small.tex
\begin{table}[t]
\centering
\scriptsize
\newcolumntype{C}{>{\centering\arraybackslash}p{2.9em}}
\begin{tabular}{r | CC | CC }

\textbf{UE} & \multicolumn{2}{c}{\textbf{\texttt{+nota}}} & \multicolumn{2}{c}{\textbf{\texttt{-correct+nota}}} \\
\textbf{Method} & AUROC & AUPRC & AUROC & AUPRC \\
\midrule
\multicolumn{4}{l}{MedGemma-4B}\\
\midrule
LabelNLL & 0.814 & 0.416 & 0.668 & 0.929  \\
ANLL & 0.662 & 0.283 & 0.607 & 0.917  \\
MaxNLL & 0.651 & 0.252 & 0.596 & 0.915  \\
SC & 0.582 & 0.230 & 0.527 & 0.885  \\
SE & 0.582 & 0.230 & 0.527 & 0.885  \\
PRO & 0.663 & 0.283 & 0.607 & 0.917  \\
EE & 0.583 & 0.217 & 0.507 & 0.897  \\
RDS & 0.592 & 0.218 & 0.498 & 0.896  \\
\textit{majority class}  & 0.500 & 0.164 & 0.500 & 0.879  \\
\midrule
\multicolumn{4}{l}{Qwen2-VL-7B}\\
\midrule
LabelNLL & 0.778 & 0.476 & 0.722 & 0.880  \\
ANLL & 0.763 & 0.432 & 0.722 & 0.878  \\
MaxNLL & 0.772 & 0.465 & 0.722 & 0.879  \\
SC & 0.767 & 0.447 & 0.686 & 0.854  \\
SE & 0.765 & 0.439 & 0.684 & 0.853  \\
PRO & 0.789 & 0.503 & 0.725 & 0.884  \\
EE & 0.749 & 0.411 & 0.680 & 0.863  \\
RDS & 0.758 & 0.431 & 0.683 & 0.865  \\
\textit{majority class}  & 0.500 & 0.236 & 0.500 & 0.777  \\
\midrule
\multicolumn{4}{l}{Qwen2.5-VL-7B}\\
\midrule
LabelNLL & 0.812 & 0.443 & 0.669 & 0.894  \\
ANLL & 0.730 & 0.357 & 0.660 & 0.898  \\
MaxNLL & 0.804 & 0.431 & 0.639 & 0.887  \\
SC & 0.773 & 0.401 & 0.632 & 0.880  \\
SE & 0.776 & 0.413 & 0.629 & 0.875  \\
PRO & 0.729 & 0.360 & 0.663 & 0.901  \\
EE & 0.707 & 0.330 & 0.600 & 0.877  \\
RDS & 0.706 & 0.324 & 0.602 & 0.878  \\
\textit{majority class}  & 0.500 & 0.176 & 0.500 & 0.839  \\
\midrule
\multicolumn{4}{l}{Molmo-7B}\\
\midrule
LabelNLL & 0.685 & 0.219 & 0.656 & 0.984  \\
ANLL & 0.555 & 0.172 & 0.527 & 0.975  \\
MaxNLL & 0.533 & 0.161 & 0.481 & 0.972  \\
SC & 0.795 & 0.396 & 0.618 & 0.980  \\
SE & 0.791 & 0.387 & 0.615 & 0.979  \\
PRO & 0.559 & 0.171 & 0.554 & 0.977  \\
EE & 0.652 & 0.257 & 0.532 & 0.977  \\
RDS & 0.655 & 0.259 & 0.529 & 0.977  \\
\textit{majority class}  & 0.500 & 0.151 & 0.500 & 0.973  \\
\midrule
\multicolumn{4}{l}{LLaVA-v1.6-7B}\\
\midrule
LabelNLL & 0.856 & 0.630 & 0.576 & 0.773  \\
ANLL & 0.769 & 0.444 & 0.564 & 0.806  \\
MaxNLL & 0.847 & 0.623 & 0.537 & 0.756  \\
SC & 0.841 & 0.566 & 0.534 & 0.769  \\
SE & 0.841 & 0.583 & 0.528 & 0.759  \\
PRO & 0.772 & 0.456 & 0.566 & 0.809  \\
EE & 0.789 & 0.493 & 0.491 & 0.749  \\
RDS & 0.797 & 0.505 & 0.494 & 0.752  \\
\textit{majority class}  & 0.500 & 0.237 & 0.500 & 0.768  \\
\bottomrule
\end{tabular}
\caption{Performance of uncertainty estimation methods in predicting correct-to-wrong answer flips for small models ($\leq$7B). \textit{majority class} rows show the baseline AUROC and AUPRC per model, accounting for class imbalances.}
\label{app_tab:nota_thresholds_appendix_small}
\end{table}

%% file: nota_thresholding_multiple_models_appendix_large.tex
\begin{table}[t]
\centering
\scriptsize
\newcolumntype{C}{>{\centering\arraybackslash}p{2.9em}}
\begin{tabular}{r | CC | CC }

\textbf{UE} & \multicolumn{2}{c}{\textbf{\texttt{+nota}}} & \multicolumn{2}{c}{\textbf{\texttt{-correct+nota}}} \\
\textbf{Method} & AUROC & AUPRC & AUROC & AUPRC \\
\midrule
\multicolumn{4}{l}{LLaVA-OV-8B}\\
\midrule
LabelNLL & 0.795 & 0.611 & 0.690 & 0.845  \\
ANLL & 0.745 & 0.498 & 0.672 & 0.852  \\
MaxNLL & 0.791 & 0.566 & 0.648 & 0.829  \\
SC & 0.777 & 0.561 & 0.653 & 0.836  \\
SE & 0.778 & 0.576 & 0.648 & 0.829  \\
PRO & 0.756 & 0.531 & 0.676 & 0.854  \\
EE & 0.723 & 0.530 & 0.607 & 0.821  \\
RDS & 0.721 & 0.523 & 0.609 & 0.823  \\
\textit{majority class}  & 0.500 & 0.309 & 0.500 & 0.780  \\
\midrule
\multicolumn{4}{l}{MedGemma-27B}\\
\midrule
LabelNLL & 0.672 & 0.369 & 0.670 & 0.875  \\
ANLL & 0.658 & 0.293 & 0.409 & 0.770  \\
MaxNLL & 0.638 & 0.271 & 0.349 & 0.717  \\
SC & 0.732 & 0.381 & 0.525 & 0.809  \\
SE & 0.731 & 0.382 & 0.525 & 0.810  \\
PRO & 0.648 & 0.281 & 0.424 & 0.777  \\
EE & 0.613 & 0.260 & 0.437 & 0.795  \\
RDS & 0.621 & 0.264 & 0.427 & 0.790  \\
\textit{majority class}  & 0.500 & 0.200 & 0.500 & 0.798  \\
\midrule
\multicolumn{4}{l}{Qwen2.5-VL-32B}\\
\midrule
LabelNLL & 0.829 & 0.523 & 0.641 & 0.756  \\
ANLL & 0.654 & 0.296 & 0.563 & 0.711  \\
MaxNLL & 0.650 & 0.280 & 0.546 & 0.702  \\
SC & 0.767 & 0.454 & 0.603 & 0.742  \\
SE & 0.767 & 0.455 & 0.602 & 0.740  \\
PRO & 0.662 & 0.302 & 0.569 & 0.715  \\
EE & 0.723 & 0.430 & 0.579 & 0.748  \\
RDS & 0.713 & 0.429 & 0.575 & 0.746  \\
\textit{majority class}  & 0.500 & 0.216 & 0.500 & 0.691  \\
\midrule
\multicolumn{4}{l}{LLaVA-v1.6-34B}\\
\midrule
LabelNLL & 0.858 & 0.536 & 0.692 & 0.855  \\
ANLL & 0.820 & 0.449 & 0.581 & 0.813  \\
MaxNLL & 0.854 & 0.534 & 0.594 & 0.833  \\
SC & 0.845 & 0.485 & 0.610 & 0.835  \\
SE & 0.843 & 0.489 & 0.605 & 0.829  \\
PRO & 0.823 & 0.460 & 0.586 & 0.818  \\
EE & 0.774 & 0.436 & 0.524 & 0.802  \\
RDS & 0.775 & 0.431 & 0.526 & 0.805  \\
\textit{majority class}  & 0.500 & 0.183 & 0.500 & 0.805  \\
\midrule
\multicolumn{4}{l}{Molmo-72B}\\
\midrule
LabelNLL & 0.872 & 0.459 & 0.770 & 0.927  \\
ANLL & 0.593 & 0.183 & 0.544 & 0.843  \\
MaxNLL & 0.494 & 0.122 & 0.519 & 0.831  \\
SC & 0.843 & 0.418 & 0.642 & 0.875  \\
SE & 0.841 & 0.422 & 0.641 & 0.875  \\
PRO & 0.593 & 0.186 & 0.551 & 0.845  \\
EE & 0.765 & 0.322 & 0.547 & 0.853  \\
RDS & 0.770 & 0.325 & 0.543 & 0.851  \\
\textit{majority class}  & 0.500 & 0.126 & 0.500 & 0.829  \\
\bottomrule
\end{tabular}
\caption{Performance of uncertainty estimation methods in predicting correct-to-wrong answer flips for medium (8B–35B) and large models ($>$35B). \textit{majority class} rows show the baseline AUROC and AUPRC per model, accounting for class imbalances.}
\label{app_tab:nota_thresholds_appendix_large}
\end{table}